\documentclass[final,3p,times,12pt]{elsarticle}

\usepackage{amssymb}
\usepackage{amsmath}
\usepackage{listings}
\usepackage{url}
\usepackage{xcolor}
\usepackage{tikz}
\usepackage{float}
\usepackage{tabularx}
\usepackage{verbatim}
\usepackage{graphicx}
\usepackage{tcolorbox}
\usepackage{pdfpages} 
\usepackage{algpseudocode}
\usepackage{algorithm}
\usepackage{pdflscape}
\usepackage{multirow}
\usepackage{dirtytalk}
\usepackage{array}
\usepackage{afterpage}
\usepackage{array,booktabs,ragged2e}

\usetikzlibrary{arrows.meta, positioning, shapes, calc} 

\usetikzlibrary{calc}
\usetikzlibrary{positioning}
\lstdefinestyle{promptstyle}{
    basicstyle=\ttfamily\small,
    backgroundcolor=\color{gray!5},
    frame=single,
    breaklines=true,
    postbreak=\mbox{\textcolor{gray}{$\hookrightarrow$}\space},
    captionpos=b,
    xleftmargin=1em,
    xrightmargin=1em,
    aboveskip=1em,
    belowskip=1em
}

\journal{Technological Forecasting and Social Change}

\begin{document}

\begin{frontmatter}



\title{Forecasting Green Skill Demand in the Automotive Industry: Evidence from Online Job Postings}


\author[label1,fn1]{Sabur Butt} 
\affiliation[label1]{organization={Institute for the Future of Education, Tecnológico de Monterrey},
            city={Monterrey},
            postcode={64849}, 
            country={Mexico}}

\author[label2,fn1]{Joshua N. Arrazola E.} 
\affiliation[label2]{organization={Universidad Politécnica de Victoria},
            city={Victoria},
            country={Mexico}}

\author[label1]{Hector G. Ceballos} 

\author[label1]{Patricia Caratozzolo} 
\fntext[fn1]{Both authors contributed equally to this research.}

\begin{abstract}
  The global transition toward sustainable economies is reshaping labor markets, yet systematic methods for identifying, forecasting, and classifying green skills remain largely undeveloped. This study presents a computational framework for measuring and forecasting green skill demand, using online job postings in Mexico's automotive industry, a sector of critical economic importance that accounts for approximately 4\% of national GDP. We construct a large-scale dataset of automotive job advertisements collected from three major Mexican recruitment platforms (Indeed México, OCC Mundial, and LinkedIn) between July 2024 and July 2025, yielding 204,373 extracted skill records. Green skills are identified through a two-stage pipeline that combines multilingual embedding similarity with validation against the ESCO (European Skills, Competences, Qualifications and Occupations), yielding 274 unique green skills across 8,576 occurrences, representing 4.22\% of all detected skills. To analyze the temporal dynamics of these competencies, we benchmark 15 state-of-the-art time series forecasting models, spanning recurrent, linear decomposition, and transformer-based architectures, using a rolling origin evaluation scheme. Transformer-based models, particularly FEDformer, Reformer, and Informer, consistently achieve the best performance, with MAE values on the order of 2.5 x $10^{-5}$ and relative RMSE values below 15. Building on these forecasts, we introduce a two-dimensional analytical framework that classifies skills by absolute and relative growth, thereby enabling the identification of stable, emerging, and high-impact competencies. The results reveal that current green skill demand in the automotive sector is concentrated in operational sustainability practices—such as logistics efficiency and environmental monitoring—while the fastest-growing competencies relate to renewable energy systems, recycling processes, and hydrogen technologies. Beyond these empirical findings, the study presents the first end-to-end pipeline that integrates skill extraction, semantic classification, time-series forecasting, and strategic skill prioritization. The proposed framework provides a scalable approach for monitoring emerging workforce capabilities and supporting data-driven workforce development policies in the context of the green transition.
\end{abstract}


\begin{highlights}
    \item First end-to-end pipeline integrating green skill extraction, semantic classification, time-series forecasting, and strategic prioritization
    \item 274 unique green skills identified from 204,373 skill records scraped from Mexican automotive job postings (July 2024--July 2025)
    \item Green skills represent 4.22\% of all detected skills, concentrated in logistics efficiency, environmental monitoring, and regulatory compliance
    \item Transformer-based models (FEDformer, Reformer, Informer) outperform recurrent and linear baselines, achieving MAE $\approx 2.5 \times 10^{-5}$
    \item A two-dimensional growth framework classifies skills into Star, Emerging, Stable, and Declining categories for workforce planning
\end{highlights}

\end{frontmatter}

\section{Introduction}
The global transition toward sustainable economies has emerged as one of the most pressing imperatives of our time, fundamentally reshaping labor markets worldwide. As nations commit to ambitious climate targets such as the Paris Agreement's goal to limit global temperature rise to 1.5°C, industries across all sectors face unprecedented pressure to adapt their operations, technologies, and workforce capabilities. This transformation has catalyzed a surge in demand for what has come to be known as ``green skills'': the knowledge, abilities, and skills necessary to develop and support sustainable, resource-efficient societies while preserving the environment for future generations \cite{buttshaping, garrote2024navigating}.

Much of the existing literature conceptualizes the green transition in terms of occupations and identifies “green jobs” or estimates employment shifts across sectors \cite{curtis2024workers, emmerling2025green}. Such approaches risk obscuring the granular dynamics through which technological change reshapes labor demand. 
The transition toward sustainability does not simply eliminate or create occupations wholesale; rather, it reconfigures the skill composition within existing roles. Engineers, logistics managers, production supervisors, and technicians increasingly require new competencies related to energy efficiency, environmental monitoring, circular production, and sustainable supply chain management \cite{atkins2024green}. Consequently, skills —not occupations— constitute the most appropriate analytical unit for understanding how decarbonization restructures labor markets \cite{fuchs2024green, zaussinger2025skills}. A skill-level perspective enables the identification of incremental transformations within occupations, the detection of emerging capabilities before occupational categories adjust, and the design of targeted training policies aligned with evolving technological trajectories \cite{li2026bridging}.


The automotive industry occupies a particularly critical position in this transformation. As one of the largest manufacturing sectors globally, contributing significantly to greenhouse gas emissions while simultaneously driving technological innovation, the automotive sector represents both a major source of environmental impact and a key lever for achieving sustainability targets. In Mexico, the automotive industry is a cornerstone of the national economy, accounting for approximately 4\% of GDP and employing over 900,000 workers directly \cite{bbva2024automotive,buttshaping}. The automotive industry is undergoing one of the most profound technological transformations of any manufacturing sector, driven by electrification, battery technologies, digitalization, and supply-chain decarbonization. These shifts fundamentally alter the skill composition required across production, logistics, and engineering functions.

Despite the critical importance of understanding how green skill requirements are evolving in Mexico's automotive sector, systematic research on this topic remains limited. Existing studies of green skills have predominantly focused on developed economies in Europe and North America \cite{garrote2024navigating, consoli2016green,buttshaping}, with comparatively little attention devoted to understanding green skill dynamics in emerging economies where manufacturing activity is increasingly concentrated. Furthermore, while considerable research has examined green skills from a static, cross-sectional perspective \cite{rutzer2020estimating}, no study so far has attempted to forecast the temporal evolution of green skill demand, a crucial gap given the rapid pace of technological and regulatory change characterizing the green transition.


Forecasting green skill demand is particularly critical because workforce development systems operate with significant time lags \cite{caratozzolo2024future}. Designing curricula, updating certification standards, allocating public training funds, and reskilling workers require months or even years before measurable labor market outcomes materialize \cite{oncioiu2025forecasting}. In rapidly evolving sectors such as automotive manufacturing, where regulatory shifts, electrification, and supply-chain decarbonization can quickly alter skill requirements, reactive policy responses risk producing structural skill mismatches. Anticipatory forecasting, therefore, becomes essential for aligning education, industrial policy, and workforce investment strategies with the projected trajectory of the green transition rather than its past manifestations \cite{thake2025transitioning}.

Addressing the green skills research gap requires innovative methodological approaches that capture both the semantic complexity of skill descriptions in job postings and the temporal dynamics of skill demand. Recent advances in natural language processing (NLP), particularly large language models such as GPT-4, offer powerful new tools for extracting structured skill information from unstructured job advertisement text~\cite{butt2025tec, azofeifa2025insights}. Concurrently, the development of sophisticated time series forecasting architectures, including transformer-based models, recurrent neural networks, and hybrid approaches, has opened new possibilities for predicting labor market trends with unprecedented accuracy \cite{garcia2022practical, chen2024job, nelson2025forecasting}.

This study makes several important contributions to the literature on green skills and labor market forecasting. First, we construct a comprehensive dataset of green skills demand in Mexico's automotive industry through large-scale web scraping of job postings from three major recruitment platforms (Indeed México, OCC Mundial, and LinkedIn) over a one-year period from July 2024 to July 2025. Second, we develop a novel methodology for identifying and classifying green skills that combines semantic embedding techniques \cite{butt2025semi,butt2025tec} with contextual validation using large language models, leveraging the ESCO (European Skills, Competences, Qualifications and Occupations) taxonomy as a reference framework \cite{esco2024classification, esco2022green}. Third, we implement and systematically compare state-of-the-art time-series forecasting models to predict the future evolution of green skill demand, including both traditional architectures (LSTM, etc.) and recent innovations (PatchTST, Crossformer, TSMixer etc.,). This comprehensive evaluation provides insights into which forecasting approaches are most suitable for capturing the complex temporal patterns characterizing green skill demand in the automotive sector. Fourth, we introduce an analytical framework for classifying skills based on their projected growth trajectories, distinguishing between ``star skills'' exhibiting high absolute and relative growth, ``emerging skills'' showing strong relative but modest absolute growth, ``stable skills'' with solid demand but limited growth potential, and ``declining skills'' facing diminishing market relevance.

The empirical findings of our analysis reveal important patterns in the evolution of green skills demand in Mexico's automotive industry. We identify 274 unique green skills across 8,576 occurrences, representing approximately 4.22\% of all detected skills. The most frequently demanded green skills relate to operational efficiency in logistics, environmental monitoring, and low-impact production processes—competencies essential for transitioning toward more sustainable automotive manufacturing practices. Our forecasting results demonstrate that transformer-based models, particularly FEDformer and Reformer, achieve superior predictive performance compared to traditional time series methods and simpler neural network architectures. These models prove especially effective at capturing the non-stationary patterns and structural breaks that characterize skill demand evolution in rapidly changing markets \cite{arenas2025demand, buttshaping}.

The structure of this paper is organized as follows. Section~\ref{sec:lr} reviews relevant literature on green skills, labor market forecasting, and the application of NLP and deep learning techniques to skill demand prediction. Section~\ref{sec:dataset} describes the dataset construction process, including data collection procedures, skill extraction, green skill identification, and time series normalization strategies. Section~\ref{sec:Methodology} presents the forecasting methodology, including the experimental setup, model specifications, and evaluation criteria. Section~\ref{sec:skill_classification_framework} introduces the analytical framework for classifying skills based on their projected growth trajectories. Section~\ref{sec:results} reports the empirical results, including model performance comparisons and skill classification outcomes. Section~\ref{sec:discusion} discusses the theoretical and practical implications of the findings, along with study limitations and directions for future research. Section~\ref{sec:conclusion} concludes the paper.

\section{Literature Review}
\label{sec:lr}
\subsection{Green Skills in the Labor Market Literature}
While the terms are sometimes used interchangeably, sustainability competences generally refer to broader cognitive and behavioral capabilities, such as systems thinking, ethical responsibility, and interdisciplinary problem solving needed to address complex sustainability challenges, whereas green skills denote more occupation and task-specific technical abilities that enable the implementation of environmentally sustainable practices within particular industries and production processes \cite{membrillo2021sustainability, montanari2023we}. Accordingly, this study adopts a green skills perspective, which enables a more precise and data-driven analysis of how environmental transitions transform the capability requirements embedded within existing occupations \cite{ibrahim2020green}. 

The concept of green skills has been widely discussed in policy and labor economics literature, yet its operational definition remains contested \cite{auktor2020green}. International organizations such as the International Labour Organization (ILO), the OECD, and CEDEFOP broadly define green skills as competencies required to support environmentally sustainable economic activities. Prior studies often distinguish between “core green skills” directly related to environmental technologies and “complementary green skills” that enable sustainability practices within existing occupations \cite{albertz2025green, outlook2023skills}. However, most empirical analyses rely on occupation-based classifications or predefined taxonomies, which may fail to capture emerging competencies reflected in real-time labor market signals such as job postings. While this approach provides useful macro-level insights, it often overlooks the fact that the majority of environmental capabilities are embedded within traditional occupations rather than entirely new job categories. As a result, recent research increasingly emphasizes the importance of analyzing green skills rather than green occupations, since environmental transitions often modify the skill composition of existing jobs rather than creating entirely new occupational categories \cite{villani2025green}.

\subsection{Computational Approaches to Skill Identification and Forecasting}
Skill forecasting has evolved from labor-intensive manual processes to sophisticated automated approaches leveraging machine learning and natural language processing \cite{rasdorf2016data, healy2015adjusting}. The advent of online recruitment platforms over the past decade has catalyzed a paradigm shift in skill demand analysis. These platforms accumulate vast volumes of job advertisement data, creating unprecedented opportunities for large-scale, real-time analysis of labor market dynamics \cite{buttshaping}. However, automated skill extraction from unstructured job postings presents significant technical challenges. Chief among these are word variability, where identical competencies are expressed using different terminology, and the continuously expanding semantic context surrounding skill descriptions \cite{butt2025tec}. For instance, skills related to \say{sustainable manufacturing} might be described variously as \say{green production,} \say{eco-friendly processes,} \say{circular economy practices,} or \say{environmental manufacturing,} complicating automated identification and classification efforts. This semantic ambiguity is particularly pronounced for emerging green skills, where terminology continues to evolve alongside technological and regulatory developments.

Recent advances in computational methods have enabled increasingly sophisticated approaches to skill demand forecasting. Early computational efforts employed statistical time series models, particularly autoregressive integrated moving average (ARIMA) models, which proved effective for certain occupational task demand predictions \cite{das2020learning, chen2024job}. Das et al. \cite{das2020learning} proposed a method for dynamic task allocation to investigate the evolution of job task requirements over a decade of AI innovation across different salary levels. However, these traditional statistical approaches often struggle to capture complex nonlinear relationships inherent in labor market dynamics and exhibit suboptimal performance when applied to large-scale datasets. We have also seen deep learning architectures such as Recurrent neural networks (RNNs), particularly Long Short-Term Memory (LSTM) networks, have demonstrated effectiveness in predicting changes in skill shares over time \cite{garcia2022practical, chen2024job}. Garcia de Macedo et al. \cite{garcia2022practical} developed practical skills demand forecasting methods via representation learning of temporal dynamics, achieving promising results in multi-step prediction tasks. However, conventional RNNs face performance degradation when handling excessively long look-back windows and forecast horizons. To address this challenge, SegRNN \cite{lin2025segrnn} introduces segment-wise iterations that reduce recurrence count within RNNs, significantly enhancing performance in time series forecasting tasks.

More recently, transformer-based models have gained widespread recognition in long-term time series forecasting due to their global modeling capabilities and attention mechanisms \cite{zhou2021informer, wu2021autoformer, zhou2022fedformer, liu2022non}. Informer \cite{zhou2021informer} incorporates low-rank matrices in self-attention mechanisms to accelerate computation for long sequence forecasting. Autoformer \cite{wu2021autoformer} employs block decomposition and autocorrelation mechanisms to more effectively capture intrinsic features of time series data. FEDformer \cite{zhou2022fedformer} utilizes DFT-based frequency-enhanced attention, while NStransformer \cite{liu2022non} addresses non-stationary time series through sequence stabilization modules. Additionally, PatchTST \cite{nie2022time} introduces channel-independent patch-based modeling that segments time series into patches, allowing the model to focus on local information and significantly improving prediction performance for skill demand forecasting \cite{chen2024job}. MLP-based models represent another promising direction, with DLinear \cite{zeng2023transformers} using series decomposition before linear regression and demonstrating that simpler architectures can sometimes outperform complex transformers in both accuracy and computational efficiency. Graph-based approaches leverage Graph Neural Networks (GNNs) to model relationships between different skills and occupations. The CHGH model \cite{chao2024cross} uses adaptive graph structures enhanced by skill co-occurrence relationships to link skill supply and demand sequences. 

Related research has emphasized the importance of understanding the labor market as a \textit{skill ecosystem}, in which competencies are interconnected through patterns of co-occurrence across occupations and industries. Studies of \textit{skill adjacency}, \textit{skill clusters}, and \textit{occupational skill networks} show that skills rarely appear in isolation but instead form structured capability bundles that shape worker mobility and technological adaptation \cite{aufiero2024mapping, liu2025patterns}. Modeling these relationships provides insights into how emerging competencies diffuse across occupations and industries, and how new technological paradigms reshape existing skill structures. Such network-based perspectives complement time-series forecasting approaches by capturing the structural relationships among skills that influence their evolution over time.

Despite rapid progress in time series forecasting architectures, most existing approaches have been developed and evaluated in contexts characterized by relatively stable or gradually evolving demand patterns. Green skill demand, however, is structurally distinct. It is shaped not only by market forces but also by regulatory interventions, technological disruptions (e.g., electrification, renewable integration), and international climate commitments \cite{knudsen2023futures}. These factors introduce structural breaks, non-stationarity, and policy-driven demand shocks that challenge conventional forecasting assumptions. Consequently, it remains unclear whether models optimized for general skill forecasting can effectively capture the unique temporal dynamics of green skill evolution \cite{caratozzolo2024future}. While we have several methods for forecasting, skill forecasting specifically lacks comprehensive, publicly accessible datasets. Recently, Chen et al. \cite{chen2024job} addressed this gap by introducing Job-SDF, a multi-granularity dataset designed for training and benchmarking job-skill demand forecasting models. Based on millions of public job advertisements collected from online recruitment platforms, Job-SDF encompasses monthly recruitment demand for 2,324 types of skills across 52 occupations, 521 companies, and 7 regions. The dataset uniquely enables evaluation of skill demand forecasting models at various granularities, including occupation, company, and regional levels. However, while Job-SDF represents an important milestone in skill forecasting research; it does not distinguish environmentally oriented competencies from general skills, nor does it provide a methodological framework for dynamically identifying green skills. As a result, the dataset enables forecasting of aggregate skill demand but does not address the sector-specific and policy-relevant challenges associated with the green transition.

Several fundamental challenges remain unaddressed in the existing literature:
\begin{itemize}
    \item There is no established framework for identifying and classifying green skills from job postings at scale. While general skill taxonomies such as O*NET \cite{levine2013net} and ESCO \cite{de2015esco} exist, the specific application of these frameworks to green skills identification requires methodological innovation, particularly given the semantic ambiguity and context-dependent nature of sustainability-related competencies. 
     \item No publicly available datasets exist that quantify green skill demand over time across different sectors, occupations, or geographic regions. 
      \item Existing skill forecasting approaches have not been systematically evaluated for their applicability to green skills, which may exhibit different temporal dynamics than general skills due to policy-driven demand shocks, technological breakthroughs in clean energy, and shifting regulatory landscapes. 
       \item There exists no pipeline or standardized methodology for continuous monitoring and prediction of green skills evolution across regions and industries.  
       \item Finally, the classification and prioritization of green skills for strategic workforce development remains an open problem. Unlike well-established frameworks for categorizing technology skills (e.g., programming languages, data analysis tools, cloud platforms), there is no consensus on how to classify green skills by growth trajectory, strategic importance, or transferability across occupations. 
\end{itemize}

Such classification schemes are critical for directing limited training resources toward skills that will have the greatest impact on both employment outcomes and environmental performance. From a labor economics perspective, the green transition can be interpreted as a directed form of skill-biased technological change, wherein environmental regulation and decarbonization targets steer innovation toward low-carbon production systems. Understanding this transformation requires moving beyond static occupation-based classifications toward dynamic, skill-level intelligence capable of anticipating emerging capability requirements. An integrated forecasting and classification framework is therefore essential for aligning workforce development systems with the evolving technological trajectory of the green economy.

The novelty of this work does not reside solely in web scraping, advanced forecasting models, or skill classification techniques taken independently. Rather, it lies in the integration of these components into the first end-to-end green skill intelligence pipeline, linking large-scale identification, temporal prediction, and strategic prioritization within a unified analytical framework. In this paper, we try to address this vacuum.

\section{Dataset}
\label{sec:dataset}

\subsection{Raw Data Collection}
The dataset used in this study was constructed through a web scraping process configured on the Apify platform~\cite{apify2025}. Job postings were collected from three of the main recruitment platforms in Mexico: Indeed México \cite{indeed2025}, OCC Mundial \cite{occ2025}, and LinkedIn \cite{linkedin2025}. Although job postings provide high-frequency signals of labor demand, they represent only a subset of the labor market and may disproportionately reflect larger firms, high-skill occupations, or digitally mediated recruitment channels \cite{magrini2023green, scoz2024mapping}. Consequently, job advertisement data should be interpreted as an indicator of formal labor demand rather than a comprehensive representation of employment dynamics \cite{fabo2022methodological}.

The extraction process focused exclusively on vacancies belonging to the Mexican automotive industry, related to seven of the most prominent companies operating in the country, namely BMW, General Motors, Toyota, Kia, Nissan, Tesla, and Volkswagen, selected based on their significant presence in the Mexican market. These firms represent the largest and most technologically advanced automotive manufacturers operating in Mexico, accounting for a substantial share of national production and employment. As leading adopters of electrification and advanced manufacturing technologies, they provide an informative lens for examining emerging green skill requirements in the sector \cite{basulto2023automotive, haar2026mexico}.

Each collected record included textual data, such as the job title, company name, full job description, and publication date. The collection period covered July 2024 to July 2025, resulting in a representative dataset of skill demand in the national automotive sector over a full year. The outcome of the data collection is shown in Table~\ref{tab:dataset_apify}. Each row represents an individual job posting obtained from one of the three previously mentioned platforms. The dataset includes the following fields: a unique identifier, the job title (\texttt{job\_name}), the full job description (\texttt{job\_description}), the publication date (\texttt{date}), and the source platform (\texttt{source}). The \texttt{job\_description} field constitutes the main source of semantic information, as it contains the complete textual content from which relevant labor skills can be identified and extracted. Once the job descriptions were collected, we conducted text cleaning and standardization. This process involved the removal of non-alphabetical characters, HTML tags, and duplicate entries to ensure textual consistency and reduce noise.

\begin{table}[]
\resizebox{\columnwidth}{!}{%
\begin{tabular}{llp{7cm}ll}
\hline
\textbf{id} & \textbf{job\_name} & \textbf{job\_description} & \textbf{date} & \textbf{source} \\ \hline
1 & Environmental Engineer &
Responsible for environmental activities in automotive manufacturing plant. Monitor emissions, waste management, compliance with environmental regulations. Work with production and maintenance teams. Knowledge of ISO standards preferred. Salary and benefits not specified. Location: Mexico. Apply online... &
2024-08-14 & Indeed \\ 
2 & Renewable Energy Technician &
Installation and maintenance of photovoltaic systems for industrial facilities. Preventive maintenance, basic diagnostics, reporting. Experience with electrical systems required. Travel may be required. Shift availability to be discussed. Compensation according to experience... &
2025-01-23 & LinkedIn \\ \hline
$\vdots$ & $\vdots$ & $\vdots$ & $\vdots$ & $\vdots$ \\ \hline
\end{tabular}%
}
\caption{Structure and representative examples of raw job descriptions collected from online job platforms using Apify.}
\label{tab:dataset_apify}
\end{table}

\subsection{Skill Extraction}

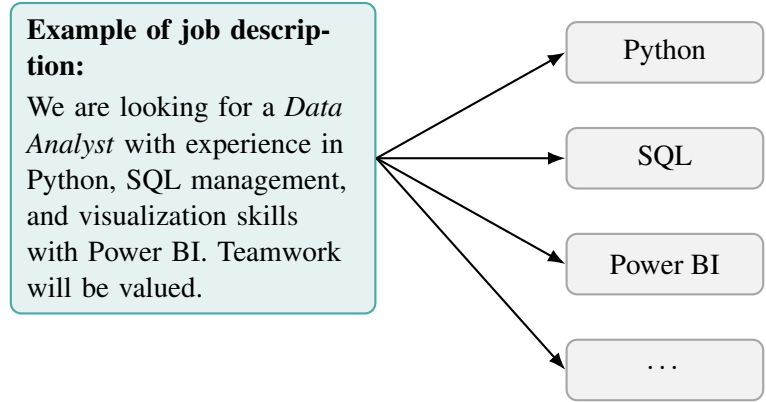
\begin{figure}[!hbpt]
\centering
\begin{tikzpicture}[
    every node/.style={font=\small},
    box/.style={
        rectangle, rounded corners,
        draw=teal!70, fill=teal!10,
        thick, minimum width=4.8cm,
        align=left, text width=4.4cm, inner sep=6pt
    },
    skill/.style={
        rectangle, rounded corners,
        draw=gray!70, fill=gray!10,
        thick, minimum width=2.6cm,
        minimum height=0.8cm,
        align=center
    },
    arrow/.style={-{Latex}, thick}
]

\node[box] (desc) {
\textbf{Example of job description:}\\[2pt]
We are looking for a \textit{Data Analyst} with experience in Python,
SQL management, and visualization skills with Power BI.
Teamwork will be valued.
};

\node[skill, right=2.5cm of desc, yshift=1.4cm] (s1) {Python};
\node[skill, right=2.5cm of desc] (s2) {SQL};
\node[skill, right=2.5cm of desc, yshift=-1.4cm] (s3) {Power BI};
\node[skill, right=2.5cm of desc, yshift=-2.8cm] (s_more) {$\cdots$};

\draw[arrow] (desc.east) -- (s1.west);
\draw[arrow] (desc.east) -- (s2.west);
\draw[arrow] (desc.east) -- (s3.west);
\draw[arrow] (desc.east) -- (s_more.west);

\end{tikzpicture}
\caption{Example of a skill extraction process from a job description using GPT-4o.}
\label{fig:skill_extraction}
\end{figure}

 After preprocessing, an initial semantic skill-extraction procedure was conducted using the GPT-4o language model, following~\citet{butt2025tec}. Butt et al. have shown the potential of skill extraction and classification in unstructured texts using LLMs, achieving promising results. Hence, we advanced on the technique, using the state-of-the-art models to identify both technical (hard skills) and interpersonal (soft skills) explicitly mentioned in each job description. If no skills were explicitly stated in the posting, the model was required to return the literal string \texttt{"Not mentioned"}. The exact structure of the prompt used for this task is presented in Appendix~\ref{app:prompt_design} (Figure~\ref{fig:prompt_skills}). This procedure generated an additional field, \texttt{Skills}, for each dataset entry, storing the extracted information as a list. Figure~\ref{fig:skill_extraction} illustrates the process, showing how relevant skills are automatically identified from the textual description of a job posting. To assess the reliability of the skill extraction procedure, a random sample of job postings was manually reviewed. The extracted skills were compared with those identified by human annotators to assess precision and recall. This validation step confirmed that the language model accurately identified explicit skill mentions while maintaining consistency across multilingual job descriptions \cite{nguyen2024rethinking}.


Since a single job advertisement may contain multiple skills expressed in either Spanish, English, or a combination of both, each record was decomposed into multiple entries, one for each detected skill. This results in a one-to-many relationship between job postings and extracted skills, where the same \texttt{Job\_ID} may appear multiple times in the dataset, each associated with a different value in the \texttt{Skills} field. Because the same skill may appear under slightly different lexical forms, a normalization step was applied to reduce redundant skill variants. Similar skill expressions were grouped using semantic similarity thresholds within the embedding space, ensuring that equivalent competencies expressed through different wording were mapped to a single skill identifier.

The presence of multilingual descriptions reflects common recruitment practices in the Mexican automotive sector. In many cases, technical terminology is retained in English while operational requirements are provided in Spanish. Additionally, some job postings are originally written in English to target bilingual candidates, resulting in non-translated skill requirements appearing directly in the description. In other cases, employers may publish duplicated versions of the same job advertisement in both languages across different platforms or within the same listing.

Consequently, the extracted skills preserve their original linguistic form, resulting in a bilingual dataset. The final structure of the dataset after this processing stage is illustrated in Table~\ref{tab:final_dataset_structure}, comprising a total of 204,373 records and five attributes.

\begin{table}[!htpb]
\centering
\renewcommand{\arraystretch}{0.1}
\setlength{\tabcolsep}{4pt}
\begin{tabular}{p{4cm} p{1.5cm} p{1.5cm} p{4.5cm} c c}
\toprule
\textbf{Title} & \textbf{Job\_ID} & \textbf{Source} & \textbf{Skills} & \textbf{Month} & \textbf{Year} \\
\midrule
asesor de servicio post venta + agencia kia 
& 0fab... 
& indeed 
& saber conducir automatico y estandar 
& 7 & 2024 \\

asesor de servicio post venta + agencia kia 
& 0fab... 
& indeed 
& conocimiento de mecanica general 
& 7 & 2024 \\

técnico en mantenimiento automotriz 
& 4a84... 
& indeed 
& conocimiento en mantenimiento preventivo y correctivo 
& 7 & 2024 \\

warranty and claims processing analyst (level 6) 
& 1c794...
& indeed 
& big data management
& 7 & 2024 \\

warranty and claims processing analyst (level 6) 
& 1c794...
& indeed 
& english level: upper intermediate 
& 7 & 2024 \\

$\vdots$ & $\vdots$ & $\vdots$ & $\vdots$ & $\vdots$ & $\vdots$ \\

fixed assets \& people cost analyst 
& aa39...
& indeed
& invoicing for miscellaneous receivables
& 7 & 2025 \\
\bottomrule
\end{tabular}
\caption{Final structure of the dataset after skill extraction. A single job posting may generate multiple records, one per detected skill, preserving the original language of the job description.}
\label{tab:final_dataset_structure}
\end{table}

\subsection{Dataset Preparation for Forecasting}
\subsubsection{Green Skills Identification and Normalization}

After extracting the set of skills from job descriptions, the next step was to determine which of these competencies could be classified as green skills, i.e., skills explicitly associated with sustainability, environmental protection, or the energy transition. To enable a systematic and standardized classification, an external reference taxonomy was required. For this purpose, the ESCO (European Skills, Competences, Qualifications and Occupations) framework was selected as the primary source of green skill definitions, owing to its explicit, structured classification of sustainability-related competencies, derived from a European Commission initiative \cite{esco2025}.

Unlike other international repositories such as O*NET, which primarily define green jobs (occupations), ESCO provides a direct classification of green skills (competencies).  
This distinction allows for a more precise alignment with the objective of this study, which focuses on identifying specific skills rather than occupational categories.

The ESCO Green Skills dataset contains a total of 386 green skills, each accompanied by a variable number of alternative labels (altLabels), ranging from one to eight. Each entry includes the following main fields:

\begin{itemize}
    \item \texttt{mainLabel}: the primary name of the skill,
    \item \texttt{altLabels}: a list of alternative names,
    \item \texttt{description}: a textual description of the skill.
\end{itemize}


To compare the skills extracted from job postings with the green skills defined in the ESCO taxonomy, both sets were embedded into a shared semantic vector space. For ESCO skills, the textual input was constructed by concatenating the main label, alternative labels, and description:

\[
\texttt{mainLabel} + \texttt{altLabels} + \texttt{description}
\]

For job-related skills, the embedding input consisted of the skill name extracted from the job description.  
All embeddings were generated using the \texttt{text-embedding-3-large} model from OpenAI \cite{openai_text_embedding_3_large}, which produces 3,072-dimensional vectors. 

An additional advantage of the proposed methodology lies in its ability to perform semantic matching across languages without requiring explicit translation. While the ESCO taxonomy is defined in English, job descriptions extracted from the Mexican labor market are mostly written in Spanish. By leveraging multilingual embeddings, the framework enables direct comparison between Spanish input text and English skill definitions within a shared semantic space. This allows queries to be executed in either language while preserving contextual meaning, reducing the need for manual preprocessing steps such as translation or dictionary-based normalization. As a result, skill identification can be performed more robustly in multilingual environments where linguistic variability would otherwise introduce additional sources of noise. Multilingual embedding models enable cross-lingual semantic alignment by mapping text from different languages into a shared vector space. In such representations, semantically equivalent phrases expressed in different languages are located near each other in the embedding space \cite{d2024alignment, kavas2025multilingual}. This property enables direct comparison of skill expressions written in Spanish with English-language skill taxonomies such as ESCO without requiring explicit translation. Recent studies in cross-lingual representation learning have demonstrated that multilingual embeddings can effectively capture semantic similarity across languages while preserving contextual meaning.

For example, the query in Spanish  \textit{"Ayudar a la reforestación"} (\textit{Help with reforestation} in English) retrieves ESCO skills related to reforestation. A sample result is shown in Figure~\ref{fig:faiss_query_example}.

\begin{figure}[!htbp]
\centering
\begin{tcolorbox}[colback=gray!4!white, colframe=teal!50!black, boxrule=0.8pt, width=0.7\textwidth]
\texttt{Query: "Ayudar a la reforestación" (Help with reforestation)} \\[4pt]
Closest results:
\begin{verbatim}
Score: 0.4316
Skill: plant trees
Text: transplant trees

Score: 0.4180
Skill: care for the wildlife
Text: flora and fauna protecting
\end{verbatim}
\end{tcolorbox}
\caption{Example of a semantic query executed in Spanish and matched against the English ESCO green skill taxonomy.}\label{fig:faiss_query_example}
\end{figure}

For each extracted job-related skill, $\text{skill}_i$, a candidate set $C_i = \{ g_1, \dots, g_5 \}$ was obtained by retrieving the five most semantically similar green skills from the ESCO taxonomy using cosine similarity as the distance metric. Retrieving the five nearest neighbors provides a balance between recall and computational efficiency while ensuring that the candidate set captures semantically related ESCO skills without introducing excessive noise. Although this procedure identifies close neighbors in the embedding space, semantic similarity alone does not guarantee conceptual equivalence. To address this limitation, each candidate set $C_i$ was evaluated using a language model (GPT-4o), which analyzed the job-related skill, its employment context, and the proposed ESCO green skills.  The model determined whether $\text{skill}_i$ could be meaningfully associated with any $g_k \in C_i$ according to the \textsc{ESCO} taxonomy. The output $y_i$ corresponded either to a valid ESCO green skill or to the special label \texttt{No}, used to discard incorrect or ambiguous associations. The Appendix~\ref{app:prompt_design} (Figure \ref{fig:prompt_green_skills}) illustrates the prompt used for this task; technical challenges related to this are presented in Appendix ~\ref{app:technical_challenges}.


The initial output of the classification pipeline followed a relational format in which each row represented a single occurrence of a skill within a specific job posting, as illustrated in Table~\ref{tab:raw_skills_format}.

\begin{table}[!htbp]
\centering
\begin{tabularx}{0.9\textwidth}{lXccc}
\hline
job\_id & esco\_green\_skill & skill\_id & month & year \\ \hline
job\_90f8d09933 & Organization and cleaning & 968 & 7 & 2024 \\ \hline
\end{tabularx}
\caption{Initial relational format of the classified skill dataset.  
Each row represents one occurrence of a skill in a job posting.}
\label{tab:raw_skills_format}
\end{table}

In this structure, the same skill could appear multiple times across different job postings and time periods. However, the forecasting models used in this study (based on the Time-Series-Library) require each variable to be represented as a single temporal sequence. Therefore, the dataset had to be transformed into a time-series matrix.

First, a frequency aggregation step was applied.  
For each skill $s$ and time period $t$ (month and year), the number of occurrences $C(s,t)$ was computed by grouping the records by $( \texttt{skill\_id}, \texttt{year}, \texttt{month})$.

Next, a pivot transformation was performed to construct a matrix in which each row corresponds to a unique skill and each column represents a monthly time period.  
Formally, the resulting matrix $\mathbf{M}$ is defined as:

\[
\mathbf{M} =
\begin{bmatrix}
C(s_1, t_1) & C(s_1, t_2) & \dots & C(s_1, t_T) \\
C(s_2, t_1) & C(s_2, t_2) & \dots & C(s_2, t_T) \\
\vdots & \vdots & \ddots & \vdots \\
C(s_n, t_1) & C(s_n, t_2) & \dots & C(s_n, t_T)
\end{bmatrix}
\]

where each row $s_i$ represents a skill and each column $t_j$ corresponds to a monthly period.  
Table~\ref{tab:pivoted_skills_format} shows the resulting structure after the pivot operation.

\begin{table}[!htbp]
\centering

\begin{tabular}{c|cccccc}
\hline
skill\_id & 2024-07 & 2024-08 & 2024-09 & 2024-10 & 2024-11 & ... \\ 
1 & 1 & 0 & 3 & 2 & 1 & ... \\ \hline
\end{tabular}
\caption{Pivoted time-series format of the skill dataset.  
Each row represents the temporal evolution of a single skill.}
\label{tab:pivoted_skills_format}
\end{table}

The final output was a matrix of size $(n_{\text{skills}}, n_{\text{periods}})$, directly suitable for time-series forecasting models.

\subsubsection{Temporal Noise and Volume Bias}
\label{subsection:volume_bias}
To avoid distortions caused by uneven data availability across time, the temporal distribution of skill occurrences was examined prior to forecasting.  
In raw monthly frequency series, patterns such as the following were observed:

\begin{center}
\texttt{89, 56, 68, 36, 37, 62, 83, 144, 92, 96, 21, 50}
\end{center}

and in other cases, even smaller and more irregular sequences:

\begin{center}
\texttt{4, 14, 14, 4, 5, 9, 10, 17, 10, 10, 2, 3}
\end{center}

These sequences exhibit abrupt increases and sharp drops that do not necessarily reflect genuine changes in skill demand. Instead, such fluctuations are largely driven by variations in the total number of job postings collected per month.  For instance, a peak of 144 occurrences in a given month does not necessarily imply a sudden increase in demand for that skill, but may simply result from a higher volume of collected vacancies or from multiple postings with similar descriptions by the same employer. To prevent the forecasting models from learning spurious patterns, it is important to account for the fact that months with more collected job postings naturally produce higher counts for all skills.  
This volume-driven effect introduces a form of structural noise, where fluctuations reflect data availability rather than genuine changes in labor market demand. Without correction, such noise could be misinterpreted by the models as meaningful trends or seasonal effects. To mitigate this bias, a temporal normalization procedure was applied. Instead of using absolute counts $C(s,t)$ for each skill $s$ in month $t$, relative proportions $\tilde{C}(s,t)$ were computed with respect to the total number of detected skills in that month.

Formally:

\begin{equation}
\tilde{C}(s, t) = \frac{C(s, t)}{N(t)}
\label{eq:normalization}
\end{equation}

where:
\begin{itemize}
  \item $C(s, t)$ denotes the number of occurrences of skill $s$ in month $t$;
  \item $N(t)$ denotes the total number of skill entries in the dataset for month $t$.
\end{itemize}

This normalization converts raw counts into comparable temporal proportions, removing the influence of sample size while preserving the relative the evolution of each skill's presence in the labor market.

\subsection{Dataset Statistics}
Table~\ref{tab:total_skills_per_month} shows the total number of detected skills in the dataset for each month.  
The values exhibit substantial variation, ranging from approximately 3,000 to more than 27,000 skill entries. This is due to multiple reasons, including seasonal job demands and platform outages related to certain platforms. However, these variations are catered to in our approach by the use of proportions.  Based on the classified dataset, all occurrences of green skills were identified for the period between July 2024 and July 2025. 
Table~\ref{tab:green_skill_counts} summarizes the monthly number of detected green skills, together with the total number of skills observed in the same period. On average, green skills account for approximately 4.22\% of all detected job-related skills, with a total of 8,576 occurrences corresponding to 274 unique green skills. Notable peaks are observed in July 2024 and March 2025.  Although an increase is observed during the first quarter of 2025, this pattern should be interpreted with caution. As we noted earlier, higher detection frequencies may reflect an increased volume of processed job advertisements rather than a structural shift in skill demand within the labor market. For this reason, the normalization procedure described in Section~\ref{subsection:volume_bias} played a central role in the analysis, as it enabled comparisons based on relative proportions rather than raw frequencies. 

\begin{table}[!htbp]
\centering
\begin{minipage}{0.35\linewidth}
    \centering
    \begin{tabular}{cc}
    \hline
    Month & Total detected skills \\ \hline
    2024-07 & 16,885 \\
    2024-08 & 15,666 \\
    2024-09 & 19,512 \\
    2024-10 & 13,166 \\
    2024-11 & 15,739 \\
    2024-12 & 16,673 \\
    2025-01 & 19,869 \\
    2025-03 & 27,289 \\
    2025-04 & 23,161 \\
    2025-05 & 24,205 \\
    2025-06 & 2,963  \\
    2025-07 & 9,245  \\ \hline
    \end{tabular}
    \caption{Monthly total number of detected skills in the dataset.}
    \label{tab:total_skills_per_month}
\end{minipage}
\hfill
\begin{minipage}{0.60\linewidth}
    \centering
    \begin{tabular}{cccc}
    \hline
    Month & Green skills & Total skills & Percentage (\%) \\ \hline
    2024-07 & 928   & 16,885 & 5.50 \\
    2024-08 & 672   & 15,666 & 4.29 \\
    2024-09 & 702   & 19,512 & 3.60 \\
    2024-10 & 354   & 13,166 & 2.69 \\
    2024-11 & 499   & 15,739 & 3.17 \\
    2024-12 & 748   & 16,673 & 4.48 \\
    2025-01 & 859   & 19,869 & 4.32 \\
    2025-03 & 1,408 & 27,289 & 5.16 \\
    2025-04 & 941   & 23,161 & 4.06 \\
    2025-05 & 920   & 24,205 & 3.80 \\
    2025-06 & 161   & 2,963  & 5.43 \\
    2025-07 & 384   & 9,245  & 4.15 \\ \hline
    \end{tabular}
    \caption{Monthly distribution of green skills and their proportion relative to the total number of detected skills.}
    \label{tab:green_skill_counts}
\end{minipage}
\end{table}

\begin{figure}[t]
\centering
\includegraphics[width=0.8\linewidth]{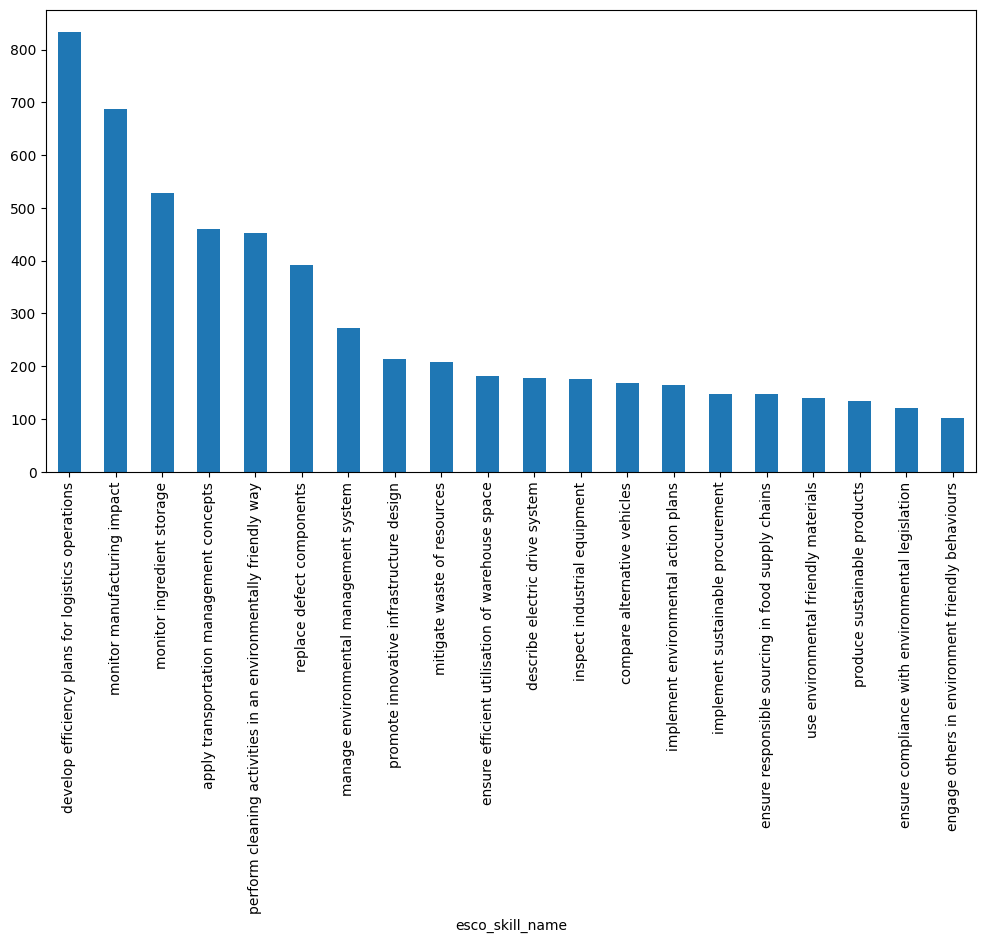}
\caption{Most frequent green skills detected in the Mexican automotive industry.}
\label{fig:top_green_skills}
\end{figure}

In addition, the frequency analysis reveals a clear \textit{long-tail} distribution, in which a small subset of skills accounts for most occurrences, while the majority appear only sporadically or at very low frequencies. This pattern is illustrated in Figure~\ref{fig:top_green_skills} and reflects a typical labor market structure, in which a limited number of competencies dominate demand, alongside a broad tail of emerging skills with lower visibility but potential for future growth. Among the most frequently observed green skills are \textit{``develop efficiency plans for logistic operations''}, 
\textit{``monitor ingredient storage''}, and \textit{``monitor manufacturing impact''}, all of which are closely related to low-impact production processes and environmental monitoring practices.  


\section{Methodology}
\label{sec:Methodology}

\subsection{Forecasting Setup}
The forecasting experiments were conducted using the models provided by the \textit{Time-Series-Library} \cite{thumlTSLib2023}, which implements a wide range of state-of-the-art architectures for time-series prediction. This study adopted a standardized experimental framework based on the Rolling Origin evaluation scheme.

Let $\{y_1, \dots, y_T\}$ denote a time series, $k$ the input sequence length (\texttt{seq\_len}), and $h$ the prediction horizon (\texttt{pred\_len}). At each time step $t$, the model $f(\cdot)$ receives the most recent $k$ observations and produces a sequence of future predictions:

\[
\widehat{\mathbf{y}}_{t+1:t+h} = f(y_{t-k+1}, \dots, y_t),
\quad \forall t \in \{k, \dots, T-h\}
\]

where $\widehat{\mathbf{y}}_{t+1:t+h} = \{\hat{y}_{t+1}, \dots, \hat{y}_{t+h}\}$ denotes the predicted values for the next $h$ time steps. The full evaluation set is defined as:

\[
\mathcal{E} =
\bigcup_{t = k}^{T-h}
\left( \widehat{\mathbf{y}}_{t+1:t+h}, \, \mathbf{y}_{t+1:t+h} \right),
\quad \text{Metric} = g(\mathcal{E})
\]

where $\mathbf{y}_{t+1:t+h}$ represents the corresponding ground-truth values and 
$g(\cdot)$ denotes the chosen performance metric.

\subsubsection{Experimental Configuration}

The models were evaluated under two temporal configurations defined by the input sequence length and the prediction horizon:

\[
k \in \{4, 6\}, \quad h = 3
\]

Thus, each model predicted the next three months using either the previous four or six months of observations.  
The Time-Series-Library includes multiple model families, which were all evaluated under the same protocol:

\begin{itemize}
    \item \textit{Recurrent and convolutional models}, designed to capture local temporal dependencies:
    \begin{center}
        \texttt{LSTM}
    \end{center}
    
    \item \textit{Decomposition and linear projection models}, aimed at separating trends, seasonality, and residual components with low computational cost:
    \begin{center}
        \texttt{TiDE, DLinear, FiLM, CHGH, TSMixer}
    \end{center}
    
    \item \textit{Transformer-based and hybrid models}, capable of modeling long-range dependencies through temporal or cross-attention mechanisms:
    \begin{center}
        \texttt{Autoformer, FEDformer, Informer, Reformer, Nonstationary\_Transformer, Transformer, Crossformer, FreTS, Koopa}
    \end{center}
\end{itemize}

\begin{figure}[H]
\centering
\begin{tikzpicture}[
    node distance=0.9cm and 0.5cm,
    every node/.style={font=\small},
    seq/.style={rectangle, rounded corners, draw=teal!70, fill=teal!10, minimum width=1cm, minimum height=0.6cm, align=center},
    pred/.style={rectangle, rounded corners, draw=orange!70, fill=orange!10, minimum width=1cm, minimum height=0.6cm, align=center},
    arrow/.style={-{Latex}, thick}
]

\node[seq] (s1) {$y_{t-3}$};
\node[seq, right=of s1] (s2) {$y_{t-2}$};
\node[seq, right=of s2] (s3) {$y_{t-1}$};
\node[seq, right=of s3] (s4) {$y_t$};

\node[pred, right=1.2cm of s4] (p1) {$\widehat{y}_{t+1}$};
\node[pred, right=of p1] (p2) {$\widehat{y}_{t+2}$};
\node[pred, right=of p2] (p3) {$\widehat{y}_{t+3}$};

\draw[arrow] (s1) -- (s2);
\draw[arrow] (s2) -- (s3);
\draw[arrow] (s3) -- (s4);
\draw[arrow, dashed] (s4) -- (p1);
\draw[arrow] (p1) -- (p2);
\draw[arrow] (p2) -- (p3);

\node[above=0.2cm of s2, teal!70!black] {$k=4$};
\node[above=0.2cm of p2, orange!70!black] {$h=3$};

\node[seq, below=1.3cm of s2] (s2b) {$y_{t-2}$};
\node[seq, right=of s2b] (s3b) {$y_{t-1}$};
\node[seq, right=of s3b] (s4b) {$y_t$};
\node[seq, right=of s4b] (s5b) {$y_{t+1}$};

\node[pred, right=1.2cm of s5b] (p6b) {$\widehat{y}_{t+2}$};
\node[pred, right=of p6b] (p7b) {$\widehat{y}_{t+3}$};
\node[pred, right=of p7b] (p8b) {$\widehat{y}_{t+4}$};

\draw[arrow] (s2b) -- (s3b);
\draw[arrow] (s3b) -- (s4b);
\draw[arrow] (s4b) -- (s5b);
\draw[arrow, dashed] (s5b) -- (p6b);
\draw[arrow] (p6b) -- (p7b);
\draw[arrow] (p7b) -- (p8b);

\node[below=0.2cm of s3b, teal!70!black] {Shifted window};

\end{tikzpicture}
\caption{Rolling Origin evaluation scheme with two consecutive sliding windows.  
Each window uses the most recent $k=4$ observations to predict the next $h=3$ time steps, and then shifts forward in time.}
\label{fig:rolling_origin_window}
\end{figure}
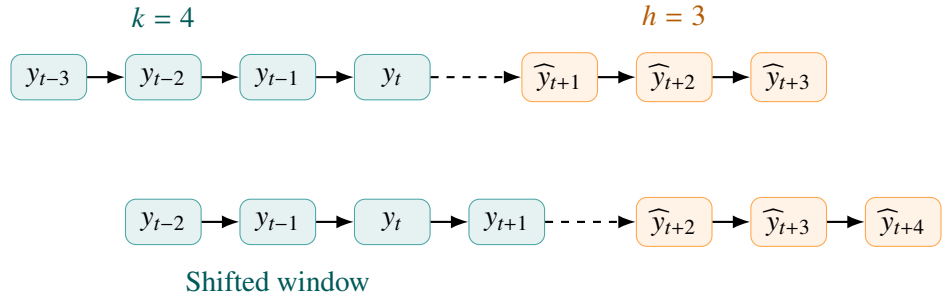

Figure~\ref{fig:rolling_origin_window} illustrates the Rolling Origin evaluation procedure using two consecutive sliding windows for $k = 4$ and $h = 3$.

The \textit{Time-Series-Library} \cite{thumlTSLib2023}, originally used for model training and evaluation, does not natively support forecasting beyond the test horizon.  To enable long-term projections, a functional extension was implemented to generate future predictions while preserving the rolling window strategy used during training. The prediction procedure follows an iterative autoregressive scheme. Starting from the last available temporal window, the trained model successively generates new prediction steps, which are appended to the input sequence and used as context for subsequent forecasts. 






\subsubsection{Hyperparameter Configuration}
To ensure fair and unbiased comparisons across different model architectures, a common set of global hyperparameters was adopted for all experiments, following the reference configuration proposed by Chen et. at. \cite{chen2024jobSDF}. This apporach avoids model-specific fine-tuning and provides a neutral baseline for performance evaluation. The selected values are reported in Table \ref{tab:model_hyperparameters}

\begin{table}[H]
\centering
\renewcommand{\arraystretch}{0.9}
\begin{tabular}{lc}
\hline
Hyperparameter & Assigned value \\ \hline
Training epochs (\texttt{train\_epochs}) & 20 \\
Batch size (\texttt{batch\_size}) & 1 \\
Early stopping patience (\texttt{patience}) & 3 \\
Learning rate (\texttt{learning\_rate}) & 0.0001 \\
Experiment description (\texttt{des}) & test \\
Loss function (\texttt{loss}) & MSE \\
Learning rate adjustment (\texttt{lradj}) & type1 \\
Model dimension (\texttt{d\_model}) & 512 \\
Number of attention heads (\texttt{n\_heads}) & 8 \\
Encoder / decoder layers (\texttt{e\_layers / d\_layers}) & 2 / 2 \\
Feed-forward dimension (\texttt{d\_ff}) & 2048 \\
Moving average window (\texttt{moving\_avg}) & 25 \\
Activation function (\texttt{activation}) & GELU \\ \hline
\end{tabular}
\caption{Hyperparameters used for the forecasting models.}
\label{tab:model_hyperparameters}
\end{table}

\subsection{Model Evaluation and Selection Criteria}
Forecasting performance was assessed using multiple complementary error metrics, capturing both absolute and relative discrepancies between predicted and observed values. Mean Absolute Error (MAE) was used to measure average absolute deviation in the original units of the time series, while Root Mean Squared Error (RMSE) emphasized larger prediction errors due to its quadratic formulation. To enable scale-independent evaluation, Symmetric Mean Absolute Percentage Error (SMAPE)  and Relative Root Mean Squared Error (RRMSE) were also computed in order to allow meaningful comparisons across time series with different magnitudes. Based on their performance, the three best-performing models were selected and subsequently used to generate 6-month future forecasts beyond the observed data horizon.

\section{Analytical Framework for Skill Classification}
\label{sec:skill_classification_framework}

Once the temporal predictions were obtained using the forecasting models described in the previous section, it became necessary to translate these numerical outputs into interpretable knowledge.  
The model outputs have limited meaning in isolation if they are not contextualized within the broader dynamics of the labor market \cite{acemoglu2022artificial}. For instance, a skill may exhibit apparent growth relative to its own historical trajectory, while still remaining marginal if other skills are expanding at a faster rate.  Conversely, a skill may experience a slight decline in frequency yet retain a dominant position within the overall skill distribution \cite{deming2020earnings}. These scenarios highlight the need for an analytical approach that combines both an individual perspective, focused on the temporal evolution of each skill, and a comparative perspective, reflecting its relative position among all skills.

To address this, a two-dimensional analytical framework was developed to evaluate the projected dynamics of skills in an integrated manner.  
This framework enables the identification of skills that show sustained growth, those that remain stable, and those that exhibit signs of decline.  
It is based on two complementary axes: absolute growth and relative growth, incorporating both historical information and model-based projections.  
Through this representation, a coherent view of green labor market dynamics can be constructed, facilitating the identification of strategic and emerging green skills \cite{knudsen2023futures}.

From a theoretical perspective, this classification framework can be interpreted as an application of capability evolution within skill ecosystems. In innovation economics and labor market studies, technological transitions rarely produce uniform changes across competencies; instead, some capabilities expand rapidly while others remain stable or decline gradually \cite{pouliakas2021understanding}. Combining absolute and relative growth allows the framework to capture both the scale of current demand and the momentum of emerging competencies. This dual perspective is particularly relevant in the context of green transitions, where new technological paradigms often coexist with established industrial practices for extended periods \cite{da2025addressing}.

Each skill $s \in \mathcal{S}$, where $\mathcal{S}$ denotes the set of all skills, is characterized using two complementary indicators:

\begin{itemize}
    \item \textbf{Axis 1: Absolute growth ($G_{\text{abs}}$)}, which reflects the change in the average occurrence level of a skill between the historical and projected periods.
    \item \textbf{Axis 2: Relative growth ($G_{\text{rel}}$)}, which measures the proportional variation with respect to the historical value, allowing skills of different magnitudes to be compared.
\end{itemize}

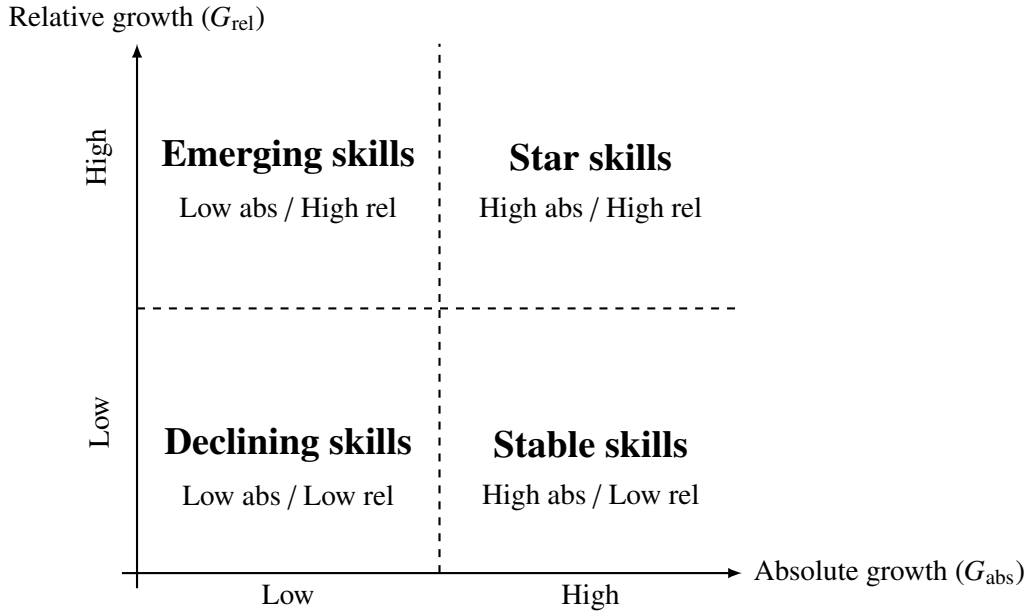
\begin{figure}[!hpbt]
\centering
\begin{tikzpicture}[>=latex, scale=1]

\draw[->, thick] (-0.2,0) -- (8,0) node[right]{\small Absolute growth ($G_{\text{abs}}$)};
\draw[->, thick] (0,-0.2) -- (0,7) node[above]{\small Relative growth ($G_{\text{rel}}$)};

\draw[dashed, thick] (4,0) -- (4,7);
\draw[dashed, thick] (0,3.5) -- (8,3.5);

\node at (6,5.5) {\large \textbf{Star skills}};
\node at (6,4.8) {\small High abs / High rel};

\node at (2,5.5) {\large \textbf{Emerging skills}};
\node at (2,4.8) {\small Low abs / High rel};

\node at (6,1.7) {\large \textbf{Stable skills}};
\node at (6,1.0) {\small High abs / Low rel};

\node at (2,1.7) {\large \textbf{Declining skills}};
\node at (2,1.0) {\small Low abs / Low rel};

\node[below] at (2,0) {\small Low};
\node[below] at (6,0) {\small High};

\node[rotate=90] at (-0.5, 2) {\small Low};
\node[rotate=90] at (-0.5, 5.5) {\small High};

\end{tikzpicture}
\caption{Skill classification based on absolute and relative growth.}
\label{fig:skill_quadrants}
\end{figure}

The resulting space is divided into four quadrants, illustrated in Figure~\ref{fig:skill_quadrants}, which distinguish between consolidated, emerging, stable, and declining skills.

For each skill $s$, the historical average over the last twelve months was computed as:

\begin{equation}
\bar{R}(s) = \frac{1}{12} \sum_{i=1}^{12} y_i(s)
\label{eq:avg_real}
\end{equation}

Similarly, the projected average over the following six months was obtained from the model forecasts:

\begin{equation}
\bar{F}(s) = \frac{1}{6} \sum_{j=1}^{6} \hat{y}_j(s)
\label{eq:avg_forecast}
\end{equation}

Absolute growth was then defined as the difference between the projected and historical averages:

\begin{equation}
G_{\text{abs}}(s) = \bar{F}(s) - \bar{R}(s)
\label{eq:abs_growth}
\end{equation}

Relative growth was computed as the ratio between absolute growth and the historical average:

\begin{equation}
G_{\text{rel}}(s) = \frac{G_{\text{abs}}(s)}{\bar{R}(s)}
\label{eq:rel_growth}
\end{equation}

Considering both absolute and relative growth is necessary because the two indicators capture distinct aspects of labor market dynamics. Absolute growth reflects the scale of demand expansion and highlights skills that are already central to industrial activity. Relative growth, in contrast, identifies competencies that are expanding rapidly from smaller baselines and may represent emerging technological capabilities. Evaluating these indicators jointly enables distinguishing between mature skills with sustained demand and nascent skills that signal structural change in production systems. Since both $G_{\text{abs}}(s)$ and $G_{\text{rel}}(s)$ are defined over the full skill set $\mathcal{S}$,  
reference thresholds were established using the 75th percentile of each distribution:

\begin{equation}
\tau_{\text{abs}} = Q_{0.75}\{\, G_{\text{abs}}(s) : s \in \mathcal{S} \,\}, \qquad
\tau_{\text{rel}} = Q_{0.75}\{\, G_{\text{rel}}(s) : s \in \mathcal{S} \,\}
\label{eq:quantile_thresholds}
\end{equation}


The use of the 75th percentile as a reference threshold provides a robust way to identify the most dynamic skills while limiting the influence of short-term fluctuations and measurement noise. Percentile-based thresholds are frequently used in labor market analytics to highlight the upper tail of growth distributions, where structural changes are most visible. By focusing on the top quartile of growth values, the framework emphasizes skills exhibiting strong momentum relative to the overall skill ecosystem. Each skill was assigned to a quadrant based on the relationship between its growth indicators and the corresponding thresholds:

\begin{equation}
\text{Category}(s) =
\begin{cases}
\text{Star}, & G_{\text{abs}}(s) \ge \tau_{\text{abs}} \;\wedge\; G_{\text{rel}}(s) \ge \tau_{\text{rel}} \\
\text{Emerging}, & G_{\text{abs}}(s) < \tau_{\text{abs}} \;\wedge\; G_{\text{rel}}(s) \ge \tau_{\text{rel}} \\
\text{Stable}, & G_{\text{abs}}(s) \ge \tau_{\text{abs}} \;\wedge\; G_{\text{rel}}(s) < \tau_{\text{rel}} \\
\text{Declining}, & G_{\text{abs}}(s) < \tau_{\text{abs}} \;\wedge\; G_{\text{rel}}(s) < \tau_{\text{rel}}
\end{cases}
\label{eq:quadrants}
\end{equation}


Beyond its analytical role, this classification framework provides a practical tool for workforce strategy and policy prioritization. Skills classified as “Star” and “Emerging” represent areas where future demand is expected to increase most rapidly and therefore constitute priority targets for training programs, curriculum development, and workforce reskilling initiatives. Conversely, “Stable” skills indicate competencies that remain essential to current industrial operations but may not require significant expansion of training capacity. In this way, the framework transforms forecasting outputs into actionable intelligence for aligning workforce development systems with the evolving requirements of the green transition.

\section{Results}
\label{sec:results}

Before discussing model performance, it is important to contextualize the forecasting task with respect to the structure of the dataset. Each skill time series comprises monthly observations over a 12-month period. While this temporal horizon is relatively short compared to typical macroeconomic forecasting datasets, the forecasting task remains meaningful because the objective is not long-term trend estimation but rather short-term demand dynamics within a rapidly evolving labor market. In addition, the analysis focuses on normalized skill proportions rather than absolute counts, thereby reducing volatility and enabling models to capture relative changes in skill demand across months. This setup aligns with recent skill demand forecasting studies using job posting data, in which high-frequency labor market signals are analyzed over relatively short time windows.

All models were evaluated under two input configurations: $\texttt{seq\_len}=4$ and $\texttt{seq\_len}=6$, while keeping the prediction horizon fixed at $\texttt{pred\_len}=3$. Table~\ref{tab:results_seq4} reports the forecasting performance for the input window ($\texttt{seq\_len}=4$), while Table~\ref{tab:results_seq6} presents the results for the longer input window ($\texttt{seq\_len}=6$).  

\begin{table}[H]
\centering
\resizebox{\columnwidth}{!}{%
\begin{tabular}{llcccc}
\toprule
\textbf{Model family} & \textbf{Model} & \textbf{MAE} & \textbf{RMSE} & \textbf{sMAPE} & \textbf{rRMSE} \\
\midrule
Recurrent / Convolutional
& LSTM & 6.49e-05 & 0.000166 & 64.70 & 30.86 \\
Decomposition / Linear
& TiDE & 8.38e-05 & 0.000204 & 71.14 & 37.96 \\
Decomposition / Linear
& DLinear & 1.03e-04 & 0.000266 & 82.74 & 49.63 \\
Decomposition / Linear
& FiLM & 8.84e-05 & 0.000201 & 68.95 & 37.53 \\
Decomposition / Linear
& CHGH & 6.68e-05 & 0.000175 & 66.58 & 32.57 \\
Decomposition / Linear
& TSMixer & 6.57e-05 & 0.000164 & 67.86 & 30.55 \\
Transformer / Hybrid
& FEDformer & \textbf{2.50e-05} & \textbf{5.92e-05} & 56.29 & \textbf{11.03} \\
Transformer / Hybrid
& Reformer & 2.54e-05 & 6.63e-05 & \textbf{55.80} & 12.36 \\
Transformer / Hybrid
& Nonstationary\_Transformer & 4.86e-05 & 0.000136 & 65.20 & 25.40 \\
Transformer / Hybrid
& Informer & 2.79e-05 & 6.33e-05 & 56.88 & 11.80 \\
Transformer / Hybrid
& Transformer & 2.61e-05 & 7.54e-05 & 56.07 & 14.05 \\
Transformer / Hybrid
& Autoformer & 3.45e-05 & 8.39e-05 & 58.35 & 15.64 \\
Transformer / Hybrid
& Crossformer & 6.47e-05 & 0.000164 & 63.45 & 30.58 \\
Transformer / Hybrid
& FreTS & 6.86e-05 & 0.000173 & 65.15 & 32.28 \\
Transformer / Hybrid
& Koopa & 8.13e-05 & 0.000219 & 72.79 & 40.82 \\
\bottomrule
\end{tabular}
}
\caption{Model performance for \(\texttt{seq\_len} = 4\), grouped by model family.}
\label{tab:results_seq4}
\end{table}

\begin{table}[H]
\centering
\resizebox{\columnwidth}{!}{%
\begin{tabular}{llcccc}
\toprule
\textbf{Model family} & \textbf{Model} & \textbf{MAE} & \textbf{RMSE} & \textbf{sMAPE} & \textbf{rRMSE} \\
\midrule
Recurrent / Convolutional
& LSTM & 6.59e-05 & 0.000168 & 64.92 & 31.21 \\
Decomposition / Linear
& TiDE & 8.95e-05 & 0.000228 & 74.33 & 42.39 \\
Decomposition / Linear
& DLinear & 8.02e-05 & 0.000214 & 71.49 & 39.83 \\
Decomposition / Linear
& FiLM & 8.74e-05 & 0.000201 & 68.06 & 37.38 \\
Decomposition / Linear
& CHGH & 6.63e-05 & 0.000175 & 66.96 & 32.53 \\
Decomposition / Linear
& TSMixer & 6.70e-05 & 0.000178 & 67.88 & 33.25 \\
Transformer / Hybrid
& FEDformer & \textbf{2.67e-05} & \textbf{6.55e-05} & \textbf{56.74} & \textbf{12.20} \\
Transformer / Hybrid
& Reformer & 3.16e-05 & 7.99e-05 & 57.38 & 14.89 \\
Transformer / Hybrid
& Nonstationary\_Transformer & 4.38e-05 & 0.000123 & 63.22 & 22.92 \\
Transformer / Hybrid
& Informer & 3.03e-05 & 7.00e-05 & 57.27 & 13.05 \\
Transformer / Hybrid
& Transformer & 3.06e-05 & 8.28e-05 & 57.19 & 15.43 \\
Transformer / Hybrid
& Autoformer & 3.83e-05 & 0.000101 & 58.90 & 18.77 \\
Transformer / Hybrid
& Crossformer & 6.16e-05 & 0.000158 & 64.12 & 29.44 \\
Transformer / Hybrid
& FreTS & 6.97e-05 & 0.000171 & 64.91 & 31.86 \\
Transformer / Hybrid
& Koopa & 7.78e-05 & 0.000212 & 72.05 & 39.42 \\
\bottomrule
\end{tabular}
}
\caption{Model performance for \(\texttt{seq\_len} = 6\), grouped by model family.}
\label{tab:results_seq6}
\end{table}

The magnitude of the reported error values should be interpreted in relation to the scale of the predicted variables. Since the forecasting models operate on normalized skill proportions rather than raw occurrence counts, the underlying values are typically very small (on the order of  \(10^{-3}\) to  \(10^{-5}\)). Consequently, absolute error metrics such as MAE and RMSE appear numerically small. However, these values represent meaningful deviations relative to the magnitude of the observed proportions. For this reason, relative metrics such as sMAPE and rRMSE provide additional insight into model performance by capturing proportional prediction accuracy across skills with different baseline frequencies.

The results show a clear and consistent advantage of transformer-based models over recurrent and linear-based approaches across both input configurations. In particular, FEDformer, Reformer, and Informer achieve the lowest error values in all evaluated metrics for \(\texttt{seq\_len}=4\) and \(\texttt{seq\_len}=6\). For instance, FEDformer attains MAE values on the order of \(2.5 \times 10^{-5}\) and RMSE values below \(6 \times 10^{-5}\) for \(\texttt{seq\_len}=4\), while maintaining similarly low errors for \(\texttt{seq\_len}=6\). Reformer and Informer follow closely, with MAE values consistently below \(3.2 \times 10^{-5}\) and rRMSE values near or below 15, indicating both high accuracy and stable behavior across configurations.
The superior performance of transformer-based architectures can be attributed to their ability to model long-range dependencies and capture complex temporal relationships through attention mechanisms. In the context of skill demand forecasting, temporal patterns may reflect non-linear dynamics driven by technological adoption cycles, regulatory changes, and fluctuations in recruitment activity. Attention-based models can dynamically weight past observations when generating predictions, enabling them to adapt more effectively to non-stationary demand patterns than recurrent or linear architectures that rely on fixed temporal dependencies.

In contrast, recurrent and linear-based models exhibit substantially higher errors. The LSTM baseline reports MAE values around \(6.5 \times 10^{-5}\) and RMSE values close to \(1.7 \times 10^{-4}\) in both configurations, roughly doubling the error observed for the best-performing transformer models. Linear-based approaches such as DLinear, FiLM, CHGH, TiDE, and TSMixer present intermediate performance, with MAE values ranging from approximately \(6.5 \times 10^{-5}\) to above \(1.0 \times 10^{-4}\), and sMAPE values frequently exceeding 65. These results reflect a more limited capacity to model complex temporal dependencies, despite the computational efficiency and structural simplicity of these architectures.

Differences across evaluation metrics further reinforce these observations. Lower RMSE values for transformer-based models indicate reduced sensitivity to large prediction errors, while the behavior under relative metrics is particularly informative in this context. Both sMAPE and rRMSE normalize the prediction error with respect to the magnitude of the observed values, enabling a fair comparison of model performance independently of scale or the number of observations. This property is especially relevant when forecasting normalized quantities such as skill proportions. Under these metrics, transformer-based models consistently outperform linear-based approaches, with sMAPE values typically around 56--58 compared to values above 65 for most linear models, and rRMSE values remaining close to or below 15, whereas linear-based models frequently exceed 30 and, in some cases, 40. This indicates that attention-based models preserve proportional accuracy more effectively across varying data scales.

Comparing the two input configurations, the \(\texttt{seq\_len}=4\) setting consistently yields slightly lower error values across most models. For example, FEDformer improves from an RMSE of \(6.55 \times 10^{-5}\) at \(\texttt{seq\_len}=6\) to \(5.92 \times 10^{-5}\) at \(\texttt{seq\_len}=4\), while similar patterns are observed for Reformer and Informer. Although these numerical differences are small in absolute terms, they are meaningful given the overall error scale, suggesting that shorter temporal windows provide a more stable representation of recent trends without introducing additional noise. These findings are consistent with trends reported in JobSDF \cite{chen2024jobSDF}, where transformer-based models generally outperform recurrent and linear baselines across multiple evaluation metrics. In both settings, attention-based architectures demonstrate greater robustness to non-stationary behavior and structural changes, while linear-based models struggle to adapt to the dynamics of the labor market.

The results indicate that forecasting skill demand proportions benefits from models that can effectively capture global temporal dependencies, using mechanisms like attention layers, while maintaining robustness to noise and non-stationarity. In this context, FEDformer, Reformer, and Informer provide the most reliable performance across configurations and metrics, justifying their selection for subsequent forecasting tasks.

After selecting the three best-performing forecasting models, the trained models were used to generate future predictions under the configuration \(\texttt{seq\_len}=4\) and \(\texttt{pred\_len}=3\). These forecasts represent the expected 6-month future evolution of ESCO green skill demand within the automotive sector. Absolute and relative growth were computed for each skill based on the predicted trajectories generated by the models. Using these values, skills were classified according to the analytical framework described in Section~\ref{sec:skill_classification_framework}. Table~\ref{tab:quadrant_distribution_models} summarizes the distribution of skills across the four quadrants for each model. In all cases, the majority of skills are classified within the \textit{Decline} quadrant, reflecting the long-tailed nature of the skill distribution, where a limited number of skills concentrate most occurrences while a large fraction exhibits low or decreasing demand.

\begin{table}[H]
\centering
\renewcommand{\arraystretch}{1.2}
\setlength{\tabcolsep}{6pt}
\begin{tabular}{|l|c|c|c|c|}
\hline
\textbf{Model} & \textbf{Star} & \textbf{Emergent} & \textbf{Stable} & \textbf{Declining} \\ \hline
Reformer & 24 & 45 & 45 & 160 \\ 
FEDformer & 33 & 36 & 36 & 169 \\ 
Informer & 27 & 42 & 42 & 163 \\ \hline
\end{tabular}
\caption{Distribution of skills classified into the four quadrants by model (\(\texttt{seq\_len}=4, \texttt{pred\_len}=3\)).}
\label{tab:quadrant_distribution_models}
\end{table}

Although the three selected models produce slightly different forecast trajectories, the resulting classification patterns remain broadly consistent. In particular, the distribution of skills across quadrants shows similar proportions of Star, Emerging, Stable, and Declining skills across models. This convergence suggests that the classification framework is relatively robust to variations in the underlying forecasting architecture. The consistency of the top-ranked skills across models further reinforces this robustness, indicating that the observed patterns reflect structural characteristics of the skill ecosystem rather than artifacts of a specific model configuration. Additionally, notable differences emerge in the allocation of skills to the \textit{Emergent} and \textit{Star} quadrants. 
FEDformer identifies a larger number of \textit{Star} skills (33 versus 24 and 27 for Reformer and Informer, respectively), whereas Reformer and Informer assign more skills to the \textit{Emergent} category. This divergence is consistent with the architectural differences among the three models, as discussed in Section~\ref{sec:discusion}.

\subsection{Absolute Growth: Established Green Competencies}

Skills with the highest absolute growth are predominantly related to logistics, operational efficiency, regulatory compliance, and process monitoring. Across all three models, \textit{develop efficiency plans for logistics operations} ranks first in absolute growth, followed by skills such as \textit{monitor ingredient storage}, \textit{apply transportation management concepts}, and \textit{mitigate waste of resources}. These skills are consistently classified within the \textit{Stable} quadrant, reflecting their sustained and substantial presence in terms of volume, even though their relative growth rate is moderate. Their dominance indicates that the current green skill landscape in the automotive sector is anchored in operational and compliance-oriented competencies, areas where sustainability intersects with established industrial practices. Complete rankings based on absolute growth are provided in Appendix~\ref{app:absolute_growth}.

\subsection{Relative Growth: Signals of an Emerging Green Transition}

In contrast, skills with the highest relative growth reveal a different and forward-looking dimension of the labor market. These skills are largely shared across models and include competencies such as \textit{monitor nature conservation}, \textit{assess hydrogen production technologies}, \textit{identify energy needs}, \textit{design geothermal energy systems}, \textit{develop recycling programs}, and \textit{inspect offshore constructions}. Nearly all of these are classified within the \textit{Emergent} quadrant, meaning they exhibit strong proportional growth from a low baseline. Their thematic concentration around energy systems, environmental monitoring, and circular economy practices suggests that the automotive sector is beginning to demand competencies that go beyond traditional operational efficiency toward more specialized sustainability knowledge. Notably, FEDformer also identifies several \textit{Star} skills with both high absolute and high relative growth, including \textit{design a domotic system in buildings}, \textit{collaborate on international energy projects}, and \textit{design heat pump installations}, pointing to a nascent but measurable demand for green engineering and energy design competencies. Full rankings based on relative growth are reported in Appendix~\ref{app:relative_growth}.

Beyond methodological considerations, the empirical results provide insights into the direction of workforce transformation within the Mexican automotive industry. The prominence of logistics optimization and environmental monitoring skills indicates that firms are prioritizing operational efficiency and regulatory compliance as immediate pathways toward sustainability. At the same time, the emergence of competencies related to hydrogen production, renewable energy systems, and recycling programs suggests that firms are beginning to invest in more transformative technologies associated with the long-term decarbonization of manufacturing systems. These patterns highlight the dual nature of the green transition, combining incremental improvements in existing processes with the gradual introduction of new technological capabilities.

From a broader perspective, the observed distinction between high-volume operational skills and rapidly growing specialized competencies reflects the layered structure of the green skill ecosystem in the automotive sector. Core operational skills related to logistics efficiency, environmental compliance, and process monitoring constitute the stable backbone of industrial sustainability practices. In parallel, emerging competencies related to renewable energy systems, hydrogen technologies, and circular economy processes represent the frontier of technological transformation. The coexistence of these two layers illustrates how the green transition unfolds incrementally within existing production systems, where new capabilities gradually complement rather than immediately replace established industrial skills.

\subsection{Cross-Model Agreement}

An important consideration is the degree to which the three models agree on skill classifications. While the exact quadrant assignments vary across models due to differences in predicted trajectories, there is substantial convergence on the key findings. The top-ranked skills by absolute growth overlap considerably: \textit{develop efficiency plans for logistics operations}, \textit{monitor ingredient storage}, and \textit{apply transportation management concepts} appear in the top five across all three models. Similarly, skills such as \textit{assess hydrogen production technologies}, \textit{monitor nature conservation}, and \textit{identify energy needs} recur in the top relative growth rankings for at least two of the three models. This consistency strengthens confidence in the robustness of the identified patterns and suggests that the classification outcomes are not artifacts of any single model's behavior. Although the three selected models exhibit broadly consistent patterns, some variation in quadrant assignments remains inevitable due to differences in their internal forecasting mechanisms. These discrepancies reflect the inherent uncertainty in predicting skill demand from short, noisy labor market time series. By comparing results across multiple high-performing models rather than relying on a single forecasting architecture, the analysis reduces the risk that classification outcomes are driven by model-specific biases. The convergence observed across FEDformer, Reformer, and Informer, therefore, provides an additional layer of robustness, suggesting that the identified emerging and dominant skills represent stable signals within the underlying labor market dynamics.

\section{Discussion}
\label{sec:discusion}

The results obtained in this study highlight the suitability of transformer-based models for forecasting skill demand in highly dynamic, nonstationary settings. Across all evaluated configurations, transformer architectures consistently outperform recurrent and linear-based alternatives, reinforcing their effectiveness when modeling fluctuating temporal patterns. However, it is important to interpret these results in light of the scale of the data. Since the analyzed skill proportions are themselves small in magnitude, even seemingly minor differences in error values represent meaningful changes in predictive performance. As such, the low absolute error values reported should not be interpreted as negligible, but rather as significant relative to the scale of the problem.

Among the evaluated models, FEDformer exhibits noticeable differences in both the number and type of skills classified as \textit{Star} and \textit{Emergent}. While all three selected models belong to the transformer family, FEDformer incorporates a mechanism based on the Discrete Fourier Transform (DFT), making it less closely aligned with the classical transformer architecture than Reformer and Informer. This architectural variation may influence how temporal patterns are captured, potentially leading to greater sensitivity to growth signals in low-frequency skills. The strong relative-metric performance observed for FEDformer, particularly under sMAPE and rRMSE, suggests that it maintains stable proportional accuracy even for less frequent skills. This property is essential when forecasting emerging competencies that are not yet dominant in the labor market. Nevertheless, these observations should be interpreted cautiously, as the available evidence does not allow for definitive attribution of causality.

The experimental results also indicate that shorter input windows tend to yield better predictive performance. While longer windows incorporate additional historical context, they may also introduce noise that degrades model generalization. This suggests that recent temporal information plays a more decisive role in short-term skill demand forecasting within the analyzed setting. These findings are broadly consistent with results reported in JobSDF \cite{chen2024jobSDF}, where transformer-based models outperform recurrent and linear baselines in skill demand prediction tasks.

\subsection{Interpreting the Green Skills Landscape}

From a broader labor economics perspective, these findings can be interpreted within the framework of skill-biased technological change. Technological transitions rarely transform labor markets uniformly; instead, they selectively increase the demand for particular capabilities while leaving others relatively unchanged. The green transition represents a directed form of technological change in which regulatory frameworks, environmental targets, and industrial innovation jointly reshape the capability requirements embedded within production systems. In this context, the coexistence of stable operational skills and rapidly growing specialized competencies reflects the gradual restructuring of industrial skill ecosystems as firms adapt to sustainability-oriented technological trajectories. The two-dimensional classification framework introduced in this study is a practical representation of capability evolution in labor markets, where absolute demand reflects the scale of existing industrial competencies and relative growth captures the momentum of emerging technological capabilities.

The skill classification results reveal a two-layer structure in the green-skills landscape of the automotive sector. The first layer consists of well-established green competencies, primarily related to logistics optimization, environmental compliance, and process monitoring. Skills such as \textit{develop efficiency plans for logistics operations} and \textit{ensure compliance with environmental legislation} dominate the \textit{Stable} quadrant across all three models. These competencies represent the current core of green workforce demand and reflect areas where sustainability requirements have already been integrated into standard industrial operations. Their high absolute presence, combined with moderate growth rates, suggests that they have reached a degree of maturity in the labor market. The second layer consists of nascent, rapidly growing competencies concentrated in the \textit{Emergent} quadrant. Skills related to hydrogen technologies, renewable energy systems, nature conservation, recycling programs, and geothermal design exhibit strong relative growth from low baselines. These skills signal the early stages of a deeper green transition, one that moves beyond operational efficiency toward specialized sustainability knowledge in energy, circular economy, and environmental stewardship. The fact that these skills are still small in absolute terms indicates that the transition is in its initial phase, but the strength and consistency of their growth trajectories across models suggest that demand is likely to intensify.

The forecasting results also illustrate the analytical value of predictive labor market intelligence. While descriptive analyses of job postings reveal current demand patterns, forecasting models allow the identification of capability signals that are only beginning to emerge. Because workforce development systems operate with significant time lags, often requiring several years to adjust curricula, training programs, and certification frameworks, the early detection of emerging skill demand is essential to align workforce preparation with future technological trajectories. In the automotive sector, the identification of rapidly growing sustainability-related skills suggests that firms are beginning to incorporate new technological competencies in energy systems, environmental monitoring, and circular production processes. Such early signals are particularly valuable for workforce development planning, as they provide advanced indications of future training needs. This layered structure suggests that green industrial transformation does not occur through the abrupt replacement of existing capabilities but through the gradual expansion of complementary skill clusters. As new technologies diffuse across production systems, emerging competencies initially appear in small numbers before progressively integrating into the broader industrial skill ecosystem.

The dominance of the \textit{Declining} quadrant, where the majority of the 274 identified green skills are classified, warrants careful interpretation. This pattern does not necessarily imply that sustainability competencies are losing relevance. Rather, it reflects the long-tailed nature of the skill distribution: a small set of core green skills accounts for the majority of demand, while a large number of specialized or niche skills appear only sporadically. This outcome is also influenced by the sparse nature of many skill time series, where low-frequency observations make it difficult to estimate stable demand trajectories within a short temporal window. Many of these low-frequency skills may be context-specific, tied to particular job roles or projects, and their classification as declining may simply reflect insufficient data to capture stable demand patterns within a twelve-month observation window. These patterns are closely aligned with the technological transformation currently underway in the global automotive industry. The transition toward electrified mobility, digital manufacturing, and decarbonized supply chains is reshaping the capability requirements of automotive firms. Traditional competencies associated with production optimization and logistics efficiency remain central to industrial operations, yet they are increasingly complemented by skills related to energy systems, environmental monitoring, and circular economy practices. The observed coexistence of these competencies suggests that the green transition in manufacturing is unfolding incrementally, with new technological capabilities gradually integrating into existing production systems rather than replacing them abruptly.

\subsection{Implications for Policy and Workforce Development}

These findings carry practical implications for several stakeholders. For policymakers and workforce development agencies, the identification of \textit{Emerging} skills provides an early signal of where training investments should be directed. For educational institutions, the results highlight a gap between the current curriculum emphasis on general industrial competencies and the emerging demand for specialized green skills. Integrating modules on renewable energy technologies, environmental auditing, sustainable procurement, and lifecycle analysis into engineering and technical programs could better prepare graduates for a labor market that is progressively valuing sustainability expertise. Similarly, for the industry, the classification framework provides a strategic tool for workforce planning. Companies can use the quadrant mapping to identify which green competencies they already possess in their workforce (\textit{Stable}), which they need to develop or recruit for (\textit{Star} and \textit{Emerging}), and which may no longer warrant significant training investment (\textit{Declining}). Beyond the empirical findings for the automotive sector, this study demonstrates the potential of integrated computational pipelines for monitoring labor market transformations in real time. By combining large-scale job posting analysis, semantic skill identification, and time-series forecasting, the proposed framework provides a scalable methodology for detecting emerging capability requirements across industries and regions. Such approaches can support the development of early-warning systems for skill shortages, enabling policymakers, educational institutions, and firms to respond more proactively to technological change.

\subsection{Methodological Contributions}

In addition to its empirical findings, this study contributes methodologically to the emerging field of computational labor market intelligence. First, it demonstrates how large-scale job posting data can be systematically transformed into structured skill demand signals through a hybrid pipeline combining semantic embeddings, taxonomy alignment, and contextual validation using large language models. This approach addresses a central challenge in labor market analytics: extracting consistent and comparable skill indicators from highly heterogeneous and multilingual job advertisement text.

Second, the study provides one of the first systematic evaluations of advanced time-series forecasting architectures for predicting skill demand in the context of the green transition. By benchmarking transformer-based, recurrent, and linear models under a unified experimental framework, the analysis identifies the forecasting approaches most capable of capturing the non-linear and non-stationary dynamics characteristic of emerging technological capabilities.

Third, integrating skill identification, forecasting, and growth-based classification within a single analytical framework is an important step toward developing scalable green skill intelligence systems. Rather than relying solely on static skill taxonomies or descriptive statistics, the proposed pipeline enables continuous monitoring of emerging capabilities and provides interpretable indicators that can support strategic workforce planning across industries and regions.

\vspace{2em}

\section{Conclusion}
\label{sec:conclusion}
This study presents a computational framework for measuring and forecasting green skill demand using large-scale online job posting data. By integrating semantic skill extraction, multilingual embedding alignment with the ESCO taxonomy, and advanced time-series forecasting architectures, the framework enables the systematic identification and prediction of sustainability-related competencies within the labor market. Applied to the Mexican automotive industry, the dataset comprises 204,373 extracted skill records, from which 274 unique green skills were identified across 8,576 occurrences, representing approximately 4.22\% of all detected skills.
The empirical analysis of Mexico’s automotive industry reveals that green skill demand currently concentrates on operational sustainability practices, particularly logistics efficiency, environmental monitoring, and compliance with environmental regulations. At the same time, forecasting results highlight the rapid emergence of competencies associated with renewable energy systems, hydrogen technologies, recycling programs, and circular production processes. These findings suggest that the green transition in manufacturing is unfolding incrementally, with new technological capabilities gradually complementing existing operational practices.
From a methodological perspective, the study demonstrates the advantages of transformer-based architectures for forecasting skill demand dynamics derived from job posting data. Models such as FEDformer, Reformer, and Informer consistently outperform recurrent and linear alternatives across multiple evaluation metrics, indicating their ability to capture non-linear and non-stationary patterns characteristic of labor market signals. Shorter input windows of four months yielded more stable predictions than longer windows, suggesting that recent temporal context is more informative than extended historical data for short-term skill demand forecasting. We further proposed a two-dimensional analytical framework that classifies skills along absolute and relative growth axes into four categories: star, emerging, stable, and declining. This classification revealed that while several green skills exhibit strong relative growth, their absolute presence in the labor market remains limited, indicating that the green transition in the automotive workforce is still in its early stages. These results contribute to the growing body of research on computational labor market intelligence by providing evidence that attention-based models are particularly well suited for high-frequency skill demand forecasting tasks. More broadly, the study demonstrates the feasibility of constructing an end-to-end green skill intelligence pipeline that integrates skill extraction, semantic classification, temporal forecasting, and strategic skill prioritization within a unified analytical framework. Beyond the automotive sector, the proposed framework offers a scalable methodology for monitoring workforce capability evolution across industries and regions. By transforming unstructured job advertisement text into structured skill time series and interpretable growth classifications, the pipeline provides actionable insights for policymakers, educational institutions, and firms seeking to anticipate emerging training needs. In this sense, the framework can support the development of early-warning systems for skill shortages and facilitate more proactive workforce development strategies aligned with technological and environmental transitions.
The study has several limitations that should be acknowledged.
First, the dataset covers only a single year of observations, which constrains the ability to capture long-term trends, seasonal effects, and structural shifts in skill demand.
Second, the analysis focuses on seven automotive companies operating in Mexico, and the findings may not fully generalize to the broader industry or to other sectors and regions.
Third, the reliance on job postings as the sole data source introduces potential biases, as not all positions are advertised online, and employer posting behavior varies across platforms and over time.
Finally, the green skill identification pipeline depends on the coverage of the ESCO taxonomy, which may not fully represent sustainability competencies specific to emerging economies or rapidly evolving technological domains. Future work could also integrate skill network analysis or skill adjacency modeling to better understand how emerging green competencies diffuse across occupations and industries. Overall, the results illustrate how computational analysis of online labor market data can contribute to understanding the evolving capability requirements of the green economy. As decarbonization policies accelerate and industrial systems undergo technological transformation, data-driven approaches for monitoring and forecasting skill demand will become essential tools for aligning workforce development systems with the evolving capability requirements of the green economy.


\bibliographystyle{ACM-Reference-Format}
\bibliography{references}
\newpage
\appendix

\section{Prompt Design}
\label{app:prompt_design}

\begin{figure}[H]
\centering
\small
\begin{tcolorbox}[colback=gray!4!white, colframe=teal!50!black, boxrule=0.8pt, width=1\textwidth]

\texttt{System:} You are an information extraction system specialized in identifying \textit{skills} within job descriptions. Your task is to recognize both technical skills (hard skills) and interpersonal skills (soft skills) explicitly mentioned in the text. \\[6pt]

\texttt{User:} The following is the job description. Extract all the skills that are explicitly mentioned and return them as a Python list.  
If no skills are mentioned, respond exactly with: "Not mentioned". \\[6pt]

\textbf{Employee Job Description:}  \begin{verbatim} [job_description] \end{verbatim}
\vspace{0.5cm}
\texttt{Developer:} You must respond strictly in the following format:

\begin{verbatim}
[skill_1, skill_2, skill_3, ...]   (if skills are mentioned)
"Not mentioned"                    (if no skills are mentioned)
\end{verbatim}
\end{tcolorbox}
\caption{Prompt used for skill extraction from job descriptions using GPT-4o.}
\label{fig:prompt_skills}
\end{figure}

\begin{figure}[H]
\centering
\small
\begin{tcolorbox}[colback=gray!4!white, colframe=teal!50!black, boxrule=0.8pt, width=1\textwidth]
\texttt{System:} You are an expert in identifying whether a skill can be considered a green skill within the ESCO taxonomy.  
A green skill is defined as the knowledge, abilities, values, and attitudes needed to live in, develop, and support a society that reduces the environmental impact of human activities.  
Evaluate whether the provided skill can be used to perform tasks that contribute to environmental sustainability. \\[6pt]
\texttt{User:} Determine whether the following skill can be semantically associated with any of the provided ESCO green skills.

Skill to classify: ``\textit{\$\{skill\_i\}}'' \\
Job context: ``\textit{\$\{job\_i\}}'' \\[4pt]
Closest green skills (candidate set $C_i = \{ g_1, \dots, g_5 \}$):
\begin{verbatim}
- {g_1}
- {g_2}
- {g_3}
- {g_4}
- {g_5}
\end{verbatim}

Decide whether the skill can be reasonably linked to any of the above based on its meaning, context, and contribution to environmental sustainability. \\[6pt]
\texttt{Developer:} You must respond strictly using the following format:
\begin{verbatim}
<Main Name>   (if a valid semantic match exists)
No            (if no match exists)
\end{verbatim}
\end{tcolorbox}

\caption{Prompt used for contextual validation of job-related skills against ESCO green skills.}
\label{fig:prompt_green_skills}
\end{figure}

\section{Technical Challenges}
\label{app:technical_challenges}
The volume of data involved in this process, both in terms of the number of job-related skills and the number of language model API calls, made sequential execution infeasible.  
An initial attempt to process the entire dataset in a single Python thread resulted in memory saturation and high request latency. To overcome these limitations, a concurrent algorithm was implemented.  
The dataset $D$ was partitioned into $N$ disjoint subsets $D_j$, each assigned to a separate thread $T_j$, such that:

    \( D = \bigcup_{j=1}^{N} D_j \quad \text{and} \quad D_i \cap D_j = \varnothing \text{ for } i \neq j \)

Each thread executed the same pipeline of neighbor retrieval, contextual validation, and partial result storage in independent files, avoiding collisions through unique identifiers. A shared global set tracked the processed $(\textit{job\_id}, \textit{skill\_id})$ pairs, ensuring consistency and uniqueness in the final mapping.

\section{Absolute Growth Rankings}
\label{app:absolute_growth}

\begin{table}[H]
  \centering
  \resizebox{\textwidth}{!}{%
  \begin{tabularx}{0.9\textwidth}{|l|X|c|c|c|}
  \hline
  \textbf{\#} & \textbf{Skill} & \textbf{$G_{\text{rel}}$} & \textbf{$G_{\text{abs}}$} & \textbf{Quadrant} \\ \hline
  1 & develop efficiency plans for logistics operations & 0.1183 & 0.000503 & Stable \\
  2 & mitigate waste of resources & 0.2759 & 0.000278 & Stable \\
  3 & monitor ingredient storage & 0.0775 & 0.000213 & Stable \\
  4 & ensure compliance with environmental legislation & 0.3852 & 0.000197 & Stable \\
  5 & apply transportation management concepts & 0.0730 & 0.000176 & Stable \\
  6 & install recycling containers & 0.6063 & 0.000132 & Star \\
  7 & ensure responsible sourcing in food supply chains & 0.1975 & 0.000123 & Stable \\
  8 & promote innovative infrastructure design & 0.1203 & 0.000122 & Stable \\
  9 & plan tanning finishing operations & 0.2671 & 0.000115 & Stable \\
  10 & carry out environmental audits & 0.9674 & 0.000107 & Star \\ \hline
  \end{tabularx}
  }
  \caption{Main green skills with greatest \textbf{$G_{\text{abs}}$} according to Reformer model.}
  \label{tab:abs_reformer}
  \end{table}
  
  \begin{table}[H]
  \centering
  \renewcommand{\arraystretch}{1.15}
  \setlength{\tabcolsep}{5pt}
  \begin{tabularx}{0.93\textwidth}{|l|X|c|c|c|}
  \hline
  \textbf{\#} & \textbf{Skill} & \textbf{$G_{\text{rel}}$} & \textbf{$G_{\text{abs}}$} & \textbf{Quadrant} \\ \hline
  1 & develop efficiency plans for logistics operations & 0.5969 & 0.002538 & Stable \\
  2 & monitor manufacturing impact & 0.3623 & 0.001189 & Stable \\
  3 & apply transportation management concepts & 0.3890 & 0.000939 & Stable \\
  4 & ensure efficient utilisation of warehouse space & 0.6634 & 0.000589 & Stable \\
  5 & monitor ingredient storage & 0.1820 & 0.000501 & Stable \\
  6 & design a domotic system in buildings & 1.7403 & 0.000368 & Star \\
  7 & collaborate on international energy projects & 1.2909 & 0.000336 & Star \\
  8 & perform cleaning activities in an environmentally friendly way & 0.1523 & 0.000329 & Stable \\
  9 & ensure correct goods labelling & 2.8806 & 0.000326 & Star \\
  10 & manage visitor flows in natural protected areas & 1.9123 & 0.000322 & Star \\ \hline
  \end{tabularx}
  \caption{Main green skills with greatest \textbf{$G_{\text{abs}}$} according to FEDformer model.}
  \label{tab:abs_fedformer}
  \end{table}
  
  \begin{table}[H]
  \centering
  \renewcommand{\arraystretch}{1.15}
  \setlength{\tabcolsep}{5pt}
  \begin{tabularx}{0.93\textwidth}{|l|X|c|c|c|}
  \hline
  \textbf{\#} & \textbf{Skill} & \textbf{$G_{\text{rel}}$} & \textbf{$G_{\text{abs}}$} & \textbf{Quadrant} \\ \hline
  1 & develop efficiency plans for logistics operations & 0.2216 & 0.000942 & Stable \\
  2 & apply transportation management concepts & 0.1820 & 0.000439 & Stable \\
  3 & monitor ingredient storage & 0.1042 & 0.000287 & Stable \\
  4 & mitigate waste of resources & 0.2570 & 0.000259 & Stable \\
  5 & ensure responsible sourcing in food supply chains & 0.3342 & 0.000209 & Stable \\
  6 & ensure compliance with environmental legislation & 0.3922 & 0.000200 & Stable \\
  7 & collaborate on international energy projects & 0.7452 & 0.000194 & Star \\
  8 & monitor manufacturing impact & 0.0438 & 0.000144 & Stable \\
  9 & implement sustainable procurement & 0.1798 & 0.000126 & Stable \\
  10 & ensure efficient utilisation of warehouse space & 0.1352 & 0.000120 & Stable \\ \hline
  \end{tabularx}
  \caption{Main green skills with greatest \textbf{$G_{\text{abs}}$} according to Informer model.}
  \label{tab:abs_informer}
  \end{table}

\section{Relative Growth Rankings}
\label{app:relative_growth}
\begin{table}[H]
  \centering
  \setlength{\tabcolsep}{5pt}
  \begin{tabularx}{0.93\textwidth}{|l|X|c|c|c|}
  \hline
  \textbf{\#} & \textbf{Skill} & \textbf{$G_{\text{rel}}$} & \textbf{$G_{\text{abs}}$} & \textbf{Quadrant} \\ \hline
  1 & measure sustainability of tourism activities & 3.0212 & 0.000009 & Emergent \\
  2 & assess hydrogen production technologies & 2.9358 & 0.000009 & Emergent \\
  3 & inspect offshore constructions & 2.4785 & 0.000008 & Emergent \\
  4 & manage plans for the utilisation of organic by-products & 2.3978 & 0.000015 & Emergent \\
  5 & monitor radiation levels & 2.3682 & 0.000009 & Emergent \\
  6 & perform selective demolition & 2.3575 & 0.000010 & Emergent \\
  7 & monitor disposal of radioactive substances & 2.3466 & 0.000007 & Emergent \\
  8 & identify energy needs & 2.2493 & 0.000009 & Emergent \\
  9 & advise on pollution prevention & 2.2342 & 0.000008 & Emergent \\
  10 & train staff to reduce food waste & 2.1014 & 0.000009 & Emergent \\ \hline
  \end{tabularx}
  \caption{Main green skills with greatest \textbf{$G_{\text{rel}}$} according to Reformer model.}
  \label{tab:rel_reformer}
  \end{table}
  
  \begin{table}[H]
  \centering
  \setlength{\tabcolsep}{5pt}
  \begin{tabularx}{0.93\textwidth}{|l|X|c|c|c|}
  \hline
  \textbf{\#} & \textbf{Skill} & \textbf{$G_{\text{rel}}$} & \textbf{$G_{\text{abs}}$} & \textbf{Quadrant} \\ \hline
  1 & dispose non-food waste within the food industry & 6.1274 & 0.000021 & Emergent \\
  2 & manage environmental impact & 5.8935 & 0.000053 & Start \\
  3 & develop natural areas works programmes & 4.4380 & 0.000125 & Start \\
  4 & measure sustainability of tourism activities & 4.1047 & 0.000013 & Emergent \\
  5 & perform selective demolition & 3.9367 & 0.000017 & Emergent \\
  6 & install heat pump & 3.7405 & 0.000062 & Start \\
  7 & monitor nature conservation & 3.3869 & 0.000021 & Emergent \\
  8 & employ habitat survey techniques & 3.3702 & 0.000055 & Start \\
  9 & design heat pump installations & 3.2952 & 0.000089 & Start \\
  10 & monitor the farm environmental management plan & 3.2379 & 0.000011 & Emergent \\ \hline
  \end{tabularx}
  \caption{Main green skills with greatest \textbf{$G_{\text{rel}}$} according to FEDformer model.}
  \label{tab:rel_fedformer}
  \end{table}
  
  \begin{table}[H]
  \centering
  \renewcommand{\arraystretch}{1}
  \setlength{\tabcolsep}{5pt}
  \begin{tabularx}{0.93\textwidth}{|l|X|c|c|c|}
  \hline
  \textbf{\#} & \textbf{Skill} & \textbf{$G_{\text{rel}}$} & \textbf{$G_{\text{abs}}$} & \textbf{Quadrant} \\ \hline
  1 & monitor disposal of radioactive substances & 2.8858 & 0.000009 & Emergent \\
  2 & monitor nature conservation & 2.3581 & 0.000014 & Emergent \\
  3 & design geothermal energy systems & 2.2811 & 0.000017 & Emergent \\
  4 & inspect offshore constructions & 2.2802 & 0.000007 & Emergent \\
  5 & develop recycling programs & 2.1972 & 0.000009 & Emergent \\
  6 & manage plans for the utilisation of organic by-products & 2.1876 & 0.000013 & Emergent \\
  7 & design drainage well systems & 2.0665 & 0.000014 & Emergent \\
  8 & assess hydrogen production technologies & 1.9418 & 0.000006 & Emergent \\
  9 & identify energy needs & 1.8860 & 0.000008 & Emergent \\
  10 & educate people about nature & 1.8812 & 0.000008 & Emergent \\ \hline
  \end{tabularx}
  \caption{Main green skills with greatest \textbf{$G_{\text{rel}}$} according to Informer model.}
  \label{tab:rel_informer}
  \end{table}


\end{document}